\documentclass{article}
\usepackage[margin=1in]{geometry} % Set 1-inch margins
\usepackage{PRIMEarxiv}
\usepackage{amsmath}
\usepackage[utf8]{inputenc} % allow utf-8 input
\usepackage[T1]{fontenc}    % use 8-bit T1 fonts
\usepackage{hyperref}       % hyperlinks
\usepackage{url}            % simple URL typesetting
\usepackage{booktabs}       % professional-quality tables
\usepackage{amsfonts}       % blackboard math symbols
\usepackage{nicefrac}       % compact symbols for 1/2, etc.
\usepackage{microtype}      % microtypography
\usepackage{lipsum}
\usepackage{fancyhdr}       % header
\usepackage{graphicx}       % graphics
\usepackage{xcolor}
\usepackage{enumitem} % Customize lists
\usepackage{adjustbox}
\usepackage{svg}
\usepackage{tabularx} % Flexible table columns
\usepackage{natbib} % Citation management
\usepackage{amsmath} % Math symbols
\usepackage{appendix}
\usepackage{subcaption}
\usepackage{threeparttable}
\usepackage{makecell}
\usepackage{array}
\usepackage{tcolorbox}
\usepackage{tabularx} % Flexible table columns
\usepackage{multirow}
\graphicspath{{media/}}     % organize your images and other figures under media/ folder
\usepackage{hyperref}
\usepackage[accsupp]{axessibility}

\usepackage{xspace}
\newcommand{\etal}{\emph{et al.}\xspace}
\newcommand{\eg}{\emph{eg.}\xspace}

\begin{document}

\title{FCMBench: The First Large-scale Financial Credit Multimodal Benchmark for Real-world Applications} 


\author{
  \textbf{Yehui Yang\textsuperscript{*}\textsuperscript{1}, 
  Dalu Yang\thanks{Equal contribution.}\ \ \textsuperscript{1}, 
  Fangxin Shang\textsuperscript{1}, Wenshuo Zhou\textsuperscript{1}, Jie Ren\textsuperscript{2}} \\
  \textbf{Yifan Liu\textsuperscript{2}, 
  Haojun Fei\thanks{Corresponding author. Contact: yangyehuisw@126.com or  haojunff@gmail.com}\ \ \textsuperscript{1}, 
  Qing Yang\textsuperscript{1}, Yanwu Xu\textsuperscript{3}\textsuperscript{,4}, Tao Chen\textsuperscript{2}} \\
  \\ 
  \textsuperscript{1} AI Lab, Qifu Technology, Beijing, China \\
  \textsuperscript{2} College of Future Information Technology, Fudan University, Shanghai, China\\
  \textsuperscript{3} School of Future Technology, South China University of Technology, Guangzhou, China \\
   \textsuperscript{4} Pazhou Lab, Guangzhou
}

\maketitle

\begin{abstract}
FCMBench is the first large-scale and privacy-compliant multimodal benchmark for real-world financial credit applications, covering tasks and robustness challenges from domain specific workflows and constraints. In this paper, we introduce FCMBench‑V1.1, which extends the language coverage of \href{https://github.com/QFIN-tech/FCMBench/tree/main/TechnicalReport}{FCMBench‑V1.0} by adding 9 English certificates and optimizes part of the instructions from the previous version. In the following description, all references to FCMBench denote FCMBench-V1.1 for simplicity.
FCMBench covers 26 certificate types, with 5,198 privacy-compliant images and 13,806 paired VQA samples. It evaluates models on \textit{Perception} and \textit{Reasoning} tasks under real-world \textit{Robustness} interferences, including 3 foundational perception tasks, 4 credit-specific reasoning tasks demanding decision-oriented visual evidence interpretation, and 10 real-world challenges for rigorous robustness stress testing. Moreover,
FCMBench offers privacy-compliant realism with minimal leakage risk through in-house scenario-aware captures of manually synthesized templates, without any publicly released images.
We conduct extensive evaluations of 28 state-of-the-art vision–language models spanning 14 AI companies and research institutes. Among them, Gemini 3 Pro achieves the best F1(\%) score as a commercial model (65.16), Kimi-K2.5 achieves the best score as an open-source baseline (60.58). The metric distribution of all tested models is $44.8\pm10.3$, indicating that FCMBench is non-trivial and provides strong resolution for separating modern vision-language model capabilities. Robustness evaluations reveal that even top-performing models experience notable performance degradation under the designed challenges. 
We have open-sourced this benchmark at \href{https://github.com/QFIN-tech/FCMBench/}{this URL} to advance AI research in the credit domain and provide a domain-specific task for real-world AI applications.
\end{abstract}

\section{Introduction}
\label{sec:intro}
Multimodal artificial intelligence (AI) has emerged as a transformative technology for financial credit businesses, where credit reviewers rely on diverse image materials uploaded by borrowers to make loan approval decisions. These workflows inherently involve two core capabilities: perception tasks (e.g., verifying the completeness of borrower-provided documents) and reasoning tasks (e.g., cross-validating stated income against bank statement deposits). Automating these tasks through AI models can significantly reduce human workload and improve loan processing efficiency—yet this progress hinges on high-quality scenario-specific benchmarks that align with real-world credit review demands. 

To better evaluate the practical applicability of models in the credit domain, a public and scientific benchmark is essential.
Existing multimodal benchmarks fail to address the unique requirements of financial credit scenarios. General-purpose benchmarks (e.g., MME \cite{fu2025mmecomprehensiveevaluationbenchmark}) assess broad perception and cognition but lack domain specificity; document understanding benchmarks (e.g., OCRBench \cite{Liu2024OCRBench}, ChartMuseum \cite{tang2025chartmuseum}, WildDoc \cite{wang2025wilddoc}) focus on OCR and chart reasoning but not credit-specific workflows; even financial-domain benchmarks (e.g., CFBenchmark-MM \cite{li2025cfbenchmarkmmchinesefinancialassistant}, FinMME \cite{luo2025finmmebenchmarkdatasetfinancial}) prioritize financial knowledge/analysis over credit review, with limited and fragmentary credit-related image and tasks due to privacy constraints and overreliance on open-source datasets. To fill this critical gap, we introduce \textbf{FCMBench}, the first multimodal benchmark explicitly tailored for financial credit—with all certificates and images independently created and captured to ensure real-world relevance and data integrity.

FCMBench comprises 13,806 vision question answering (VQA) samples, covering 26 certificate types (see Figure~\ref{fig:certificatesOverview}). To comply with privacy regulations, all personal information (e.g., persona settings, institution/location names, logos, and seals) is fully fictionalized. Our team manually designed material templates, produced physical certificates, and conducted on-site photography to ensure data authenticity and uniqueness.

This work makes three-fold key contributions:
\begin{enumerate}
\item{\textbf{Filling the gap in credit-specific multimodal benchmarks.}}
To resolve the privacy-driven unavailability of shareable real credit data, we release the first large-scale credit-focused multimodal benchmark. All physical certificates and images in FCMBench are self-created and never publicly available before, addressing the lack of scenario-specific data for credit AI evaluation.

\item{\textbf{Innovating an application-oriented evaluation system.}}
We design an evaluation framework aligned with real-world credit review workflows, integrating \textit{Perception} (3 tasks) and \textit{Reasoning} (4 tasks) capabilities across 10 \textit{Robustness} challenges. This system enables quantitative evaluation of models' practical capabilities—from information understanding to risk judgment—under realistic application conditions.

\item{\textbf{Promoting academia-industry collaboration in the credit domain.}}
By open-sourcing FCMBench, we resolve two industry dilemmas: financial institutions gain a standard for comparing credit AI models, while researchers (academia and fintech) access high-quality data for in-depth studies. This breaks down data barriers and drives credit AI from single-point optimization to collaborative innovation.
\end{enumerate}

\section{Related Work}
Multimodal benchmarks have shifted from general-domain to scenario-specific evaluations. We review relevant existing multimodal benchmarks as follows.
%\paragraph{General Multimodal Benchmarks}

\subsubsection{General Multimodal Benchmarks}
General multimodal benchmarks assess Multimodal Large Language Model(MLLMs) in domain-agnostic scenarios to evaluate comprehensive capabilities. Yue \etal proposed MMMU \cite{yue2024MMMU} for college-level multidisciplinary knowledge and reasoning; its extension MMMU-Pro \cite{yue2025mmmupro} optimized evaluation by filtering text-answerable questions and introducing a vision-only setting, resolving early visual content redundancy. Chen \etal \cite{chen2024mmstar} addressed data leakage and redundant visuals—key flaws in existing evaluations—via 1,500 vision-indispensable samples and multimodal gain metrics. Fu \etal \cite{fu2025mmecomprehensiveevaluationbenchmark} proposed MME, the first comprehensive MLLM benchmark covering 14 perception and cognition subtasks, with manually designed instruction-answer pairs ensuring fairness and avoiding data leakage.

\subsubsection{Document Understanding Benchmarks}
Document understanding benchmarks focus on MLLMs' processing of structured/unstructured document images for OCR, chart reasoning, etc. Liu \etal \cite{Liu2024OCRBench} developed OCRBench, the most comprehensive OCR-focused benchmark covering 29 datasets for text recognition, document VQA, and KIE, identifying model weaknesses in multilingual, handwritten, and non-semantic text. Tang \etal \cite{tang2025chartmuseum} and Masry \etal \cite{masry2025chartqaprodiversechallengingbenchmark} proposed ChartMuseum and ChartQAPro respectively: the former revealed a large human-model gap (93\% vs. 63.0\% top model accuracy) and poor visual reasoning, while the latter addressed real-world diversity shortages in existing chart benchmarks. Zhu \etal \cite{zhu2024mmdocbenchbenchmark} introduced MMDocBench for fine-grained document perception and reasoning, and Wang \etal \cite{wang2025wilddoc} proposed WildDoc—the first natural-environment document benchmark—exposing state-of-the-art MLLMs' poor robustness under variable illumination and distortions.

\subsubsection{Financial-domain Multimodal Benchmarks}
Financial multimodal benchmarks, a specialized subset of document understanding, focus on financial texts, figures, and tables. Early Chinese-centric efforts include CFBenchmark-MM \cite{li2025cfbenchmarkmmchinesefinancialassistant} with over 9,000 image-question pairs for financial explanation and knowledge, and its extension \cite{liu2025visfinevalscenariodrivenchinesemultimodal} expanding to 15,848 pairs covering data analysis, asset optimization, etc. Luo \etal \cite{luo2025finmmebenchmarkdatasetfinancial} addressed the lack of specialized financial benchmarks with 11,000 samples across 18 domains (e.g., Technology, Consumer, Pharmaceuticals), while Deng \etal \cite{deng2025finmrknowledgeintensivemultimodalbenchmark} focused on expert-level reasoning with 3,200 annotated pairs in investment, valuation, and risk models. Recent advances include MultiFinBen \cite{peng2025multifinbenbenchmarkinglargelanguage} pioneering multilingual and multimodal (text, vision, audio) evaluation with financial OCR tasks, and Yang \etal \cite{yanglet2025multimodalfinancialfoundationmodels} providing a conceptual framework summarizing progress, challenges, and prospects of financial multimodal foundation models.

\begin{figure}[htb]
\centering
\vspace{-0.5em} % 小幅减小图片顶部间距（可选，进一步腾空间）
\includegraphics[width=0.9\linewidth]{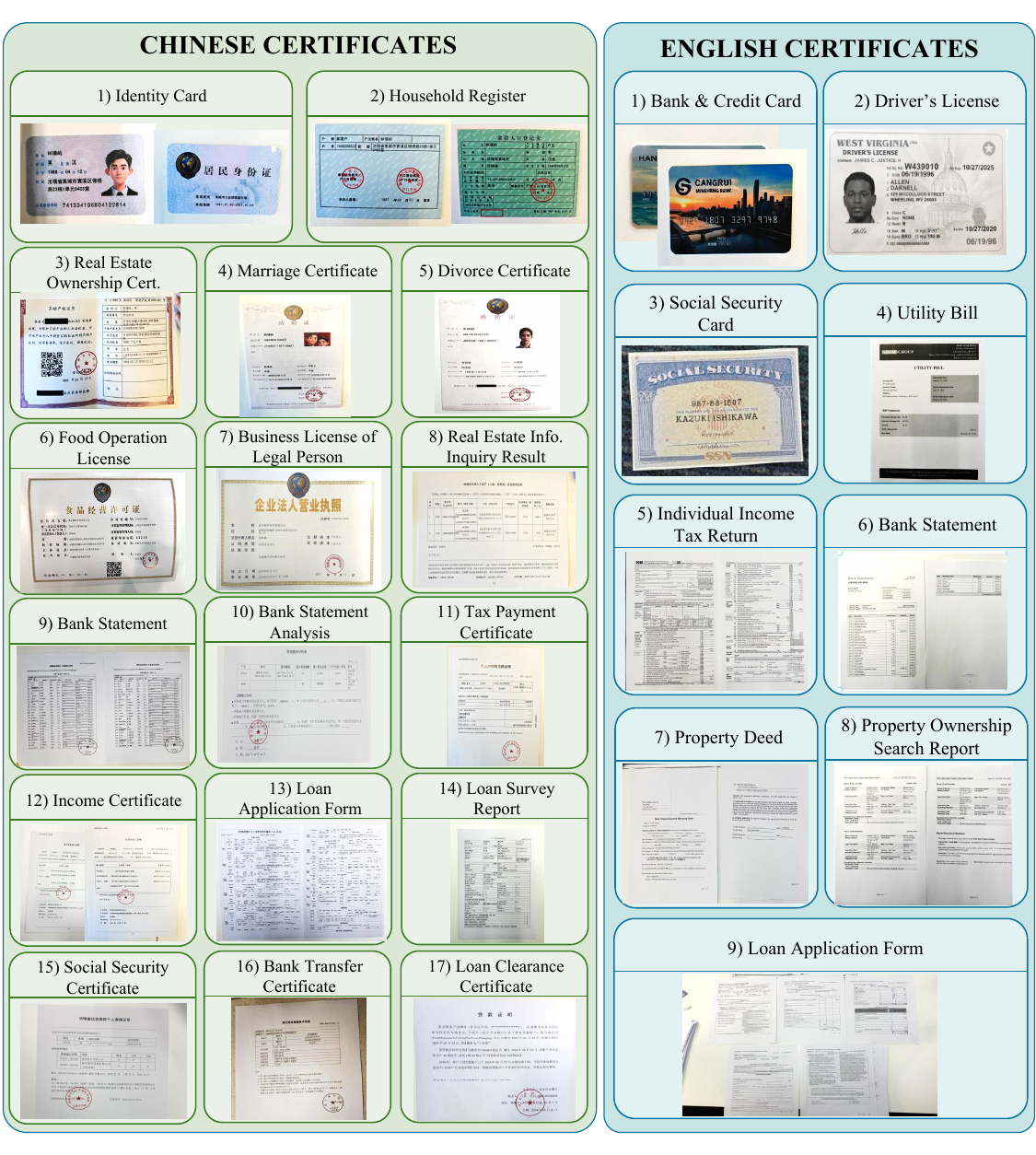}
\caption{Overview of the 26 categories of certificates in FCMBench.}
\vspace{-0.3em} % 小幅减小图片底部间距（可选）
\label{fig:certificatesOverview}
\end{figure}

\section{Benchmark Construction}
\subsection{Overview and Statistics}
FCMBench comprises 26 categories of certificates spanning both Chinese and English documents commonly used in credit review (Figure~\ref{fig:certificatesOverview}). The benchmark is guided by two design principles: (1)\textit{ alignment with real credit review workflows} and (2)\textit{ diversity of data forms}. Specifically, the certificate categories and task designs are grounded in interviews with over 20 senior credit reviewers from commercial banks and financial credit companies, ensuring practical relevance across the full review chain (e.g., loan review, income verification, and asset evaluation). Moreover, FCMBench incorporates heterogeneous document forms from standardized cards and certificates to materials containing complex charts and tables, and supports both single-image and multi-image inputs, reflecting the variety of evidence encountered in real applications. To better approximate real submissions, each image can be matched to a complete set of an individual’s materials rather than being randomly collected.

\begin{figure}[h]
\centering
\includegraphics[width=1.0\linewidth]{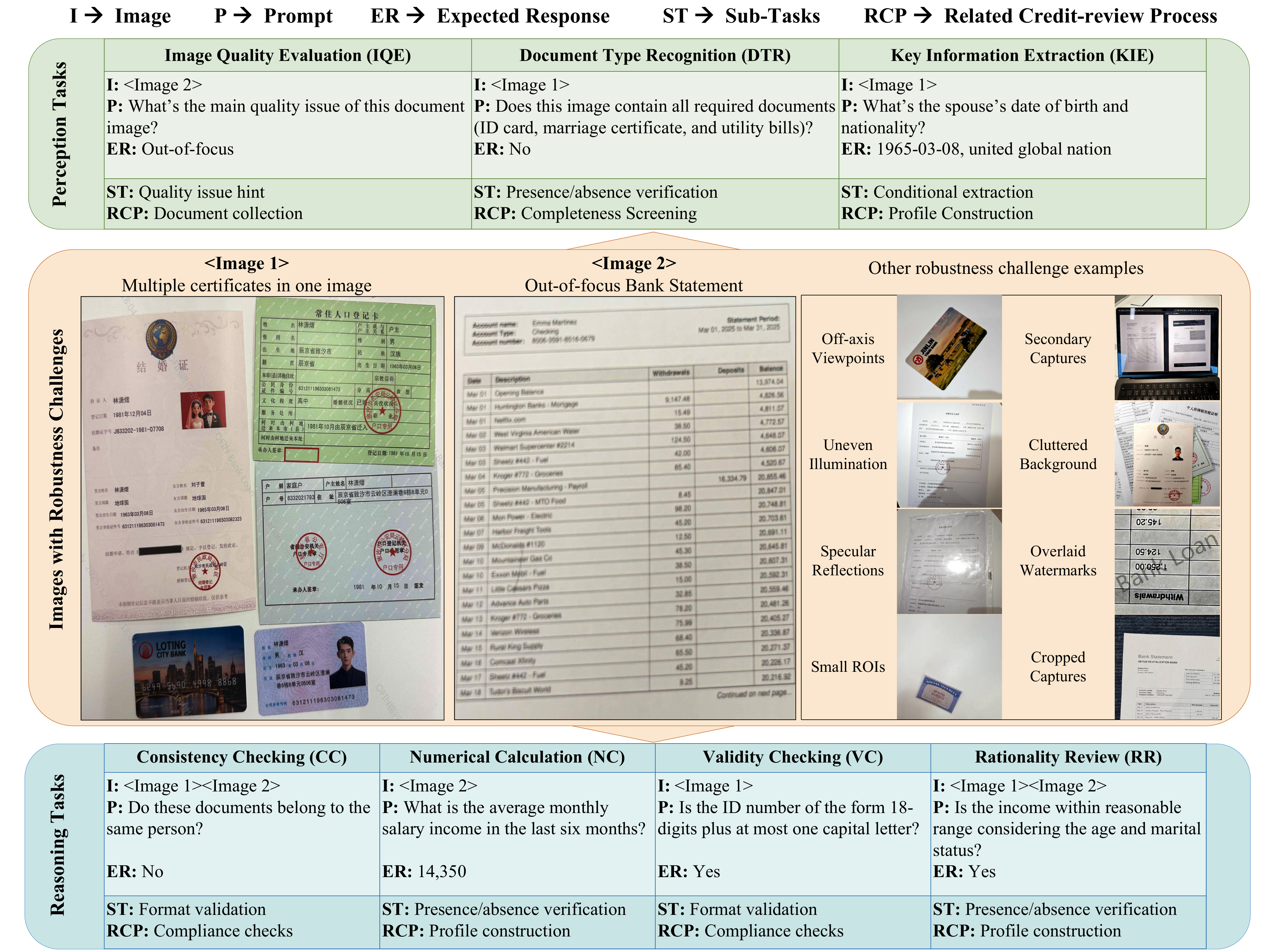}
\caption{Top and bottom: subtask examples of perception and reasoning tasks, showing input image, core prompt (omitting options and format requirements), expected response, and related real-world financial credit review workflows. Middle: example input images and snippets of different types and robustness challenges.}
\label{fig:task_challenges_framework}
\end{figure}

To evaluate model capability under practical conditions, FCMBench is organized into perception, reasoning, and robustness settings. As illustrated in Figure~\ref{fig:task_challenges_framework}, perception tasks assess the extraction of visual and textual information from certificate images, while reasoning tasks evaluate credit-related decision making grounded in the extracted evidence. Robustness challenges capture common real-world interferences. Task definitions and challenge statistics are summarized in Table~\ref{tab:perception_reasoning_tasks} and Table~\ref{tab:challenge_statistics}. Since a single image may be associated with multiple tasks and challenges, the aggregated image counts in these tables do not equal the number of unique images.

\begin{table}[h]
\centering
%\scriptsize
\small
\caption{Taxonomy and Statistics of \textit{Perception} and \textit{Reasoning} Tasks. DTR: Document Type Recognition; KIE: Key Information Extraction; IQE: Image Quality Evaluation; CC: Consistency Checking; VC: Validity Checking; NC: Numerical Calculation; RR: Rationality Review}
\label{tab:perception_reasoning_tasks}
\begin{tabularx}{\linewidth}{lXcc}
\toprule
Task\qquad \qquad  & Description & \# Images & \# QA pairs \\
\midrule
IQE & To recognize quality issues (\eg, specular reflection, out-of-focus blur) on the certificates in the image. & 830 & 830 \\
\addlinespace
DTR & To recognize the certificates contained in the images. An image may include one or more certificates. & 3908 & 5030 \\
\addlinespace
KIE & To extract key information or key-value pairs in the certificate from the given images & 3360 & 3327 \\
\bottomrule
\addlinespace
CC & To check whether the documents in an image set belong to the same person, or to cross-check a listing document has its corresponding supporting document. & 2216 & 1239\\
\addlinespace
VC & To check whether documents have valid values, \eg, document has not expired, values follow required formats. & 1400 & 1442 \\
\addlinespace
NC & To compute numbers based on the information presented in the images. & 1105 &  1614 \\
\addlinespace
RR & To check if the values presented in different documents are within reasonable range, \eg, income vs tax certificate, income vs bank statement. & 515 & 324 \\
\bottomrule
\end{tabularx}
\end{table}

\begin{table}[h]
\centering
%\scriptsize
\small
\caption{Taxonomy and Statistics of Normal Captures and \textit{Robustness} Challenges}
\label{tab:challenge_statistics}
\begin{tabularx}{\linewidth}{lXcc}
\toprule
Challenge Name & Description & \# Images & \# QA pairs \\
\midrule
Normal Captures & A "standard" photo of the certificates that satisfies all quality requirements. & 568 & 1219 \\
\addlinespace
Off-axis Viewpoints & The certificates are captured from rotated viewpoints. & 577 & 1306\\
\addlinespace
Uneven Illumination & Various shadows cast onto the certificates. & 556 & 1301\\
\addlinespace
Specular Reflections & Over-exposition brought by the light. & 581 & 1272\\ 
\addlinespace
Out-of-focus & Blur caused by defocus. & 556 & 1306\\
\addlinespace
Small ROIs & The target certificate only takes a small position (<20\%) in the image. & 553 & 1223\\
\addlinespace
Secondary Captures & Photos are captured from other on-screen images (\eg, computer displays) rather than the original certificates. & 512 & 891\\
\addlinespace
Cluttered Background & The backgrounds of the target certificates share similar characteristics, which may affect the recognition of the certificates. & 540 & 1162\\
\addlinespace
Overlaid Watermarks & Watermarks on the image or on the certificates may affect the recognition. & 552 & 1081 \\
\addlinespace
Cropped Captures & The certificates shown in the image are incomplete. & 528 & 853\\
\addlinespace
Multi-doc Images & Models must perform tasks involving multiple certificates captured in a single image. & 774 & 2192 \\
\bottomrule
\end{tabularx}
\end{table}

\subsection{Image Generation}
To mitigate privacy risks and ensure data compliance, all images in FCMBench are generated via a fully controlled synthetic-to-physical workflow, which comprises three sequential stages with rigorous standardization.

%\begin{enumerate}
\subsubsection{1. Certificate Template Construction:}
\begin{figure}[htb]
    \centering
    \includegraphics[width=1.0\linewidth]{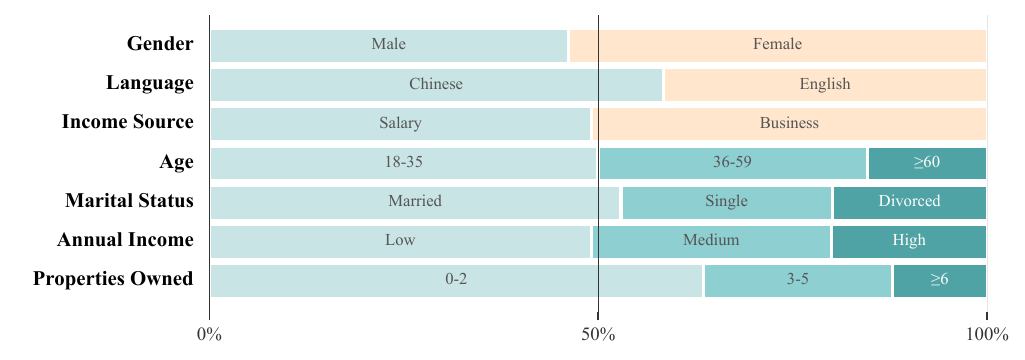}
    \caption{Demographic status distribution of the fictional identities in FCMBench}
    \label{fig:character_distributions}
\end{figure}

\begin{figure}[ht]
    \centering
    \includegraphics[width=\linewidth]{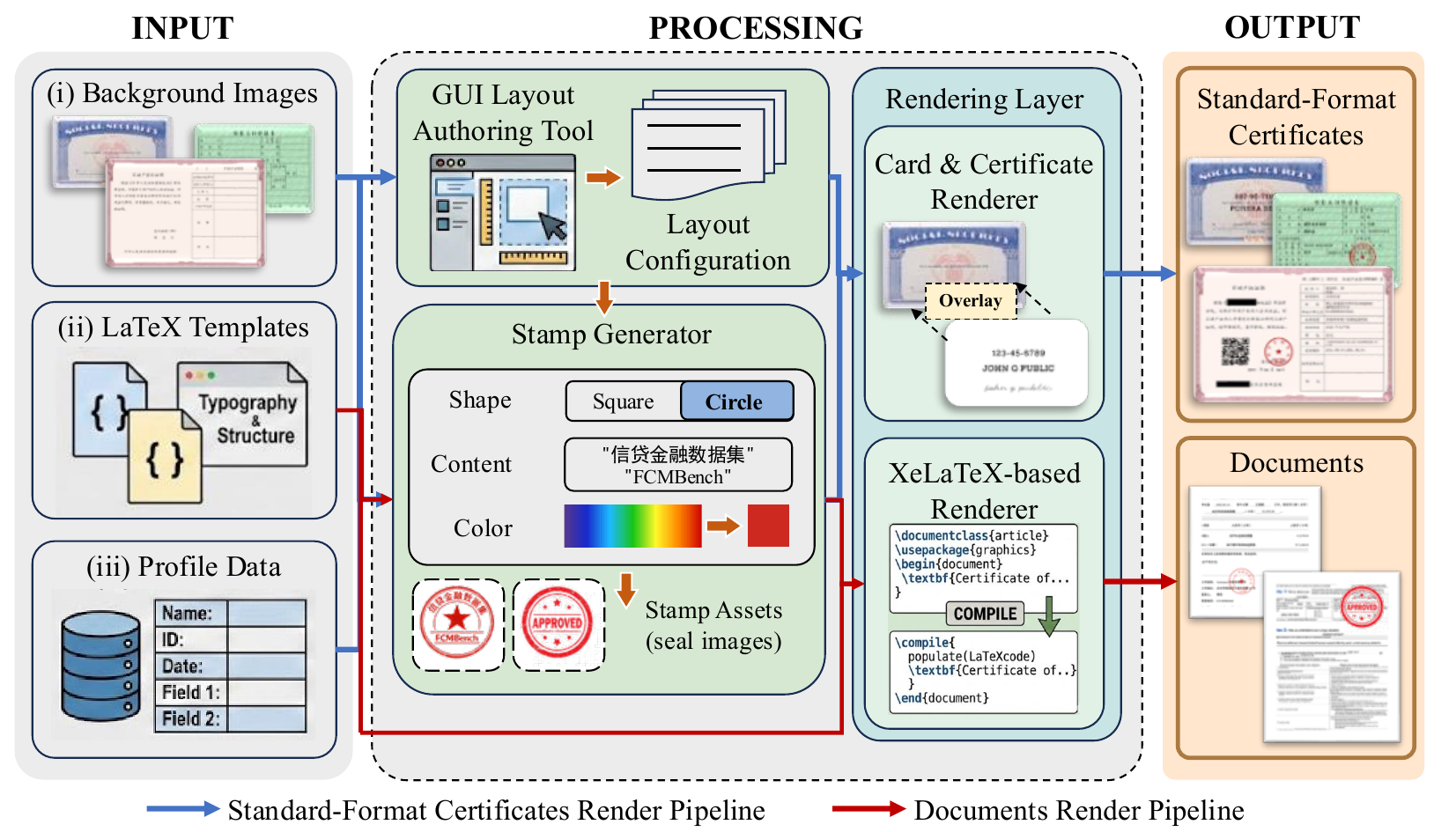}
    \caption{Overview of the synthetic credit-document generation toolchain in FCMBench, which combines a card and certificate renderer and a LaTeX--YAML compiler to generate template-based synthetic credit documents that are printed and re-captured as realistic benchmark images.}
    \label{fig:qfin_synthesis_pipeline}
\end{figure}

We develop a high-fidelity synthetic certificate generation pipeline to produce credit-related documents and cards that are closely analogous to real-world counterparts. The detailed synthesis workflow is illustrated in Figure~\ref{fig:qfin_synthesis_pipeline}. Specifically, FCMBench is organized around a moderate-sized pool of fictional identities with diverse demographic and economic attributes (Figure~\ref{fig:character_distributions}). For each identity in the pool, we generate a complete application profile spanning multiple document types, and physically recapture the resulting documents to simulate real-world acquisition. This controlled design enables cross-document consistency and traceability (e.g., matching names, IDs, addresses, and financial attributes across certificates), while keeping other factors (template, capture conditions, and perturbations) systematically measurable.  

\subsubsection{2. Compliance Integration:}
To eradicate real-world identifiers and comply with data protection regulations, with the exception of a small subset (15 English-language driver’s licenses) adopted from the synthetic IDNet dataset~\cite{xie2024idnet}, most logos, institutional emblems, and portraits are replaced with AIGC-synthesized alternatives. Additionally, a fictional institutional ecosystem was constructed, including simulated streets, government agencies, and banking entities. All tasks in FCMBench are anchored to the same set of fictional identities to ensure identity consistency across certificates and reasoning scenarios.

\subsubsection{3. Physical Fabrication and Scene-based Shooting:}
Following the synthesis of electronic certificate templates, all certificates—including driver’s licenses from the IDNet dataset—were physically fabricated: either as high-fidelity card replicas or printed on standard A4 paper, ensuring dimensional, layout, and material properties consistent with their real-world equivalents. A team of 11 participants takes images of these physical documents under ecologically valid capture conditions (detailed in Table~\ref{tab:challenge_statistics}). %Images were captured using five major smartphone brands (iPhone, Huawei, Honor, Xiaomi, and Oppo) to account for real-world device heterogeneity and typical acquisition challenges.

\subsection{Instruction Design}\label{subsec:instruction-design}
As presented in Fig.~\ref{fig:task_challenges_framework}, all tasks in FCMBench are abstracted from real-world credit review workflow, ensuring strong practical relevance—each task directly mirrors concrete decisions and operations performed by credit review officers or automated systems in actual scenarios. Below, we detail the design of perception and reasoning tasks, each explicitly mapped to the core credit review workflow to reflect real application requirements: \emph{Document Collection}, \emph{Completeness Screening}, \emph{Profile Construction}, \emph{Cross-Document Consistency Checks}, \emph{Compliance Checks}, and \emph{Income Rationality and Risk Assessment}.

\subsubsection{Perception Task Design}
Perception tasks are designed to replicate the front-end information acquisition and preliminary screening stages of credit review.% directly corresponding to the \emph{Document collection}, \emph{Completeness screening}, and \emph{Profile construction} procedures.
\begin{enumerate}
\item{Image Quality Evaluation (IQE):}
Aligned with \emph{Document collection} procedure, IQE models the ``desk-reject'' stage for low-quality submissions. It requires selecting the dominant issue (or ``no issue'') from common acquisition defects (consistent with the challenge types in Table~\ref{tab:challenge_statistics}).

\item{Document Type Recognition (DTR):}
Aligned with \emph{Completeness screening} procedure, DTR evaluates whether a model can identify document categories (single or multi-image) and detect missing mandatory document types required by a given product/policy. We include prompts for single-image classification, multi-image aggregation, and presence/absence screening, mirroring real completeness checks.

\item{Key Information Extraction (KIE):}
Aligned with \emph{Profile construction} procedure, KIE measures a model’s ability to extract key fields and values (e.g., identity, income, property attributes) needed to populate internal profiles and application forms. Prompt variants cover key-presence checks, value extraction, and conditional extraction.

\end{enumerate}

\subsubsection{Reasoning Task Design}
Reasoning tasks correspond to the middle and back-end decision-making stages of credit review, replicating the complex analytical and verification workflows performed by credit officers.% and are mapped to \emph{Cross-document consistency checks}, \emph{Compliance checks}, and \emph{Income rationality and risk assessment} procedure.

\begin{enumerate}
\item{Consistency Checking (CC).}
Aligned with \emph{Cross-document consistency checks} procedure, CC evaluates whether information is coherent across multiple documents, covering identity matching (\eg, \emph{Identity Card} vs.\ \emph{Household Register}), document linkage (\eg, \emph{Real Estate Inquiry} vs.\ \emph{Ownership Certificate}), and transaction reconciliation (\eg, \emph{Receipt} vs.\ \emph{Bank Statement}) to surface mismatches and missing linkages.

\item{Validity Checking (VC).}
Aligned with \emph{Compliance checks} procedure, VC tests whether documents satisfy explicit policy rules, including issue/expiry date validation, format requirements, and basic authenticity cues, mirroring routine compliance screening in credit review.

\item{Numerical Calculation (NC).}
Aligned with \emph{Profile construction}, \emph{Income rationality} and \emph{Risk assessment} procedure, NC measures quantitative reasoning over documents, requiring aggregation and transformation of numeric evidence (\eg, totals, differences, and derived indicators) for income verification and collateral coverage checks.

\item{Rationality Review (RR).}
Aligned with \emph{Income rationality} and \emph{ Risk assessment} procedure, RR assesses plausibility-based ``sanity checks'' by verifying whether values are mutually consistent and economically reasonable across documents (\eg, \emph{Income Certificate} with \emph{Tax Certificate}, \emph{Loan Application Form}, or \emph{Bank Statement}).

\end{enumerate}

\subsubsection{Robustness Integration}
All tasks are instantiated on both normal and low-quality captures (off-axis viewpoints, uneven illumination, specular reflection, etc., inherited from the photo collection. See Table~\ref{tab:challenge_statistics}) for challenge description and middle part of Figure~\ref{fig:task_challenges_framework} for image examples. %In some low-quality captures, even humans may find it hard to complete the task. This may cause an overall underestimation of model performance in terms of the absolute values, but will serve a fair comparison across all models.

\subsection{Evaluation Metrics}\label{{subsec:metric}}
In credit review scenarios (\eg loan application), many fields (\eg, ID numbers, addresses, bank account identifiers and dates) are legally and operationally atomic: partially correct or semantically equivalent paraphrases are not acceptable. Therefore, FCMBench adopts an exact-match oriented evaluation for atomic values. We intentionally do not treat semantic equivalence (e.g., paraphrased addresses or reworded entities) as correct, because such outputs are not actionable in downstream compliance checks and risk control workflows. To reduce ambiguity, we explicitly specify output formats (\eg, date and currency representations) in the prompts, and evaluate model outputs under these requirements.

We flatten diverse output formats into order-invariant key-value sets \(S = \{(k_i, v_i)\}_i\) with tailored normalization:
\begin{itemize}    
\item[-]\textit{Dictionaries}: Add hierarchical key prefixes (e.g., \texttt{"field.subfield"}).
\item[-]\textit{List-of-lists}: Convert inner lists to tuples, normalize elements, map to a common prefix.
\item[-]\textit{Primitive values}: Normalize strings (trim whitespace/brackets, split on commas) and numbers (e.g., \texttt{"12.0"}=\texttt{"12"}); NC tasks tolerate numerical differences <2.
\end{itemize} 
Set-based precision (P), recall (R), and F1 are calculated as:
\[\text{P} = \frac{|S_{\text{pred}} \cap S_{\text{gt}}|}{|S_{\text{pred}}|}, \quad\text{R} = \frac{|S_{\text{pred}} \cap S_{\text{gt}}|}{|S_{\text{gt}}|}, \quad\text{F1} = \frac{2 \cdot \text{P} \cdot \text{R}}{\text{P} + \text{R}}.\]
Empty \(S_{\text{gt}}\) yields F1=1 if \(S_{\text{pred}}=\emptyset\) (else 0); F1 is reported as 0–100.

The final F1 score is aggregated across three levels:
\begin{enumerate}
\item{Instance level}: F1=0–1 (multi-valued outputs) or 0/1 (simple strings/choices).
\item{Subtask level}: Average F1 for instances in the same fine-grained subtask.
\item{Task level}: Macro-average subtask scores for each of the seven tasks.
\end{enumerate} 

This design unifies diverse output formats (labels, lists, tables, numerics) for cross-model comparison; macro-averaging avoids over-reliance on single patterns, emphasizing stable credit-document skill performance. 

For robustness challenges, we compute F1 scores in the same way but report them by image artifact type. For each artifact, we normalize its F1 by the F1 on Normal Captures, yielding a relative F1 ratio that quantifies robustness degradation (or retention) under that specific artifact.

\section{Experiments}
\subsection{Experimental Setup}
We evaluate 28 SOTA VLMs released in 2025--2026 by 14 AI companies and research institutes, spanning commercial and open-source models across diverse scales (Table~\ref{tab:sota_mmmodels}). Detailed settings are listed below:

\subsubsection{Deployment Settings}
Commercial models and $>$300B models are accessed via API with temperature $0.01$ and images resized to the API limit (other parameters default). All other models are deployed on Alibaba Cloud PAI (NVIDIA H20, 96\,GB HBM3) using ms-swift~\cite{msswift2024}, with topK, topP, temperature, max model length, and max new tokens set to 1, 0.001, 0, 31744, and 1024, respectively, and images are resized to $\sim$6M pixels. All models use the same prompt and test samples.
Note that DeepSeek-OCR \cite{deepseekocr} has recently achieved impressive performance across various document OCR tasks, which motivated us to investigate its performance on our FCMBench. Since DeepSeek-OCR only supports a subset of perception tasks, we concatenate it with DeepSeek V3.2 \cite{deepseek3.2}, treating this two-stage pipeline as a unified VLM to ensure a fair and consistent comparison with other VLMs.

\subsubsection{Reasoning effort settings}
Since our evaluation focuses on practical extraction and compliance-checking performance under tight latency constraints, we prefer to adopt model \emph{Instruct} variants where available, or set reasoning controls to the lowest-effort mode (e.g., \emph{none}/\emph{minimum}) in API requests. 

\begin{table}[ht]
\centering
\caption{Overview of SOTA VLMs}
\label{tab:sota_mmmodels}
\begin{threeparttable}
\small
%\footnotesize
%\scriptsize
%\tiny
\begin{tabularx}{\linewidth}{l X X X} % 从6列改为5列（去掉Deployment）
\toprule
\textbf{Model} & \textbf{Developer} & \textbf{Model Size} & \textbf{Release Date} \\
% \multirow{2}{*}{\textbf{Model}} &
% \multirow{2}{*}{\textbf{Developer}} &
% \multirow{2}{*}{\makecell[c]{\textbf{Model}\\\textbf{Size*}}} &
% \multirow{2}{*}{\makecell[c]{\textbf{Reasoning}\\\textbf{Mode}\tnote{**}}} &
% \multirow{2}{*}{\makecell[c]{\textbf{Release}\\\textbf{Date}}} \\
%\multirow{2}{*}{\textbf{Deployment}} \\
% & & & \\
\midrule
\multicolumn{4}{l}{\textbf{Commercial Models}} \\ % 修正为跨5列
\addlinespace
Claude Opus 4.5 \cite{claude_opus} &  Anthropic &  Not disclosed  &  Nov-25 \\%Think  &  Nov-25 \\
Gemini 3 Pro/Flash  \cite{gemini3}  & Google DeepMind  & Not disclosed  & Nov/Dec-25 \\% Think  &  Nov/Dec-25 \\
GPT 5.1/5.2 \cite{gpt5_1}  & OpenAI  & Not disclosed  & Nov/Dec-25\\% Think  &  Nov/Dec-25\\
% Grok 4  \cite{grok4}  & xAI  & /  &  Think  &  Jul-25 \\
\midrule
\multicolumn{4}{l}{\textbf{Open-Source VLMs}} \\ % 修正为跨5列
\addlinespace
Qwen3.5 Series\cite{qwen3.5} & Alibaba Cloud & 397B/A17B & Feb-26 \\%None & Feb-26 \\
Kimi-K2.5 \cite{kimik2.5}  & Moonshot AI  &  1T/A32B  & Feb-26 \\%  Off  &  Feb-26 \\
Qwen3-VL Series\cite{qwen3vl}  & Alibaba Cloud & 8B-235B/A22B   & Oct-25 \\%  None  &  Oct-25 \\
LLaVA-OneVision-1.5 \cite{llavaonevision}  & LMMS Lab & 8B  & Sep-25 \\% None  &  Sep-25 \\
InternVL-3.5 Series\cite{internvl35}  & Shanghai AI Lab  &  8B-241B/A28B & Aug-25 \\% None  &  Aug-25 \\
Ovis2.5 \cite{ovis25}  & Alibaba (AIDC-AI)  &  9B  & Aug-25\\% Off  &  Aug-25 \\
MiniCPM-V 4.5 \cite{minicpmv45}  & ModelBest & 8B  & Aug-25\\% Think  &  Aug-25 \\
GLM-4.5/4.6V \cite{glm45v}  & Zhipu AI  & 106B/A12B  & Jul/Dec-25 \\% Off  &  Jul/Dec-25 \\
Kimi-VL \cite{kimivl}  & Moonshot AI  &  16B/A3B  & Jun-25 \\% None  &  Jun-25 \\
Llama-4-Maverick \cite{llama4}  & Meta AI  & 400B/A17B  & Apr-25 \\% Think  &  Apr-25 \\
Phi-4-Multimodal \cite{phi4mm}  & Microsoft  & 6B  & Feb-25 \\% None  &  Feb-25 \\
Minimax-01 \cite{minimaxvl}  & MiniMax AI  & 456B/A46B  & Jan-25 \\% None  &  Jan-25 \\
Janus-Pro \cite{januspro}  & DeepSeek AI  & 7B  & Jan-25 \\% None  &  Jan-25 \\
%\midrule
%\multicolumn{4}{l}{\textbf{OCR + LLM}} \\ % 修正为跨5列
%\addlinespace
DeepSeek(OCR\cite{deepseekocr}+V3.2\cite{deepseek3.2}) & DeepSeek AI & 3B + 671B & Sep/Dec-25\\% None & Sep/Dec-25 \\
% \makecell[l]{DeepSeek-OCR \cite{deepseekocr} \\ \qquad \quad + \\ DeepSeek V3.2 \cite{deepseek3.2}} & DeepSeek AI & 3B + 671B & None & Sep-25 \\
\bottomrule
\end{tabularx}
% \begin{tablenotes}
% \scriptsize
% \item [*] "/": the total parameter count is not publicly disclosed. "xB-xxB": multiple released sizes within the same model family. "AxxB": active parameter counts in an Mixture-of-Experts model. \\
% \item [**] Reasoning Mode denotes how we handle the model's chain-of-thought ability for testing. "None": the model does not explicitly support reasoning. "Off": the model accepts a switch variable to turn chain-of-thought off. "Think": the chain-of-thought behavior cannot be turned off, and the answer can be inspired by thinking. 
% \end{tablenotes}
\end{threeparttable}
\end{table}

\subsection{Experimental Results and Analysis}
\subsubsection{Overall Performance}

We conduct a multi-dimensional evaluation of SOTA VLMs on FCMBench to assess both benchmark difficulty and discriminability. 

\begin{figure}[htb]
    \centering
    \begin{subfigure}[t]{0.48\linewidth}
        \includegraphics[width=\linewidth]{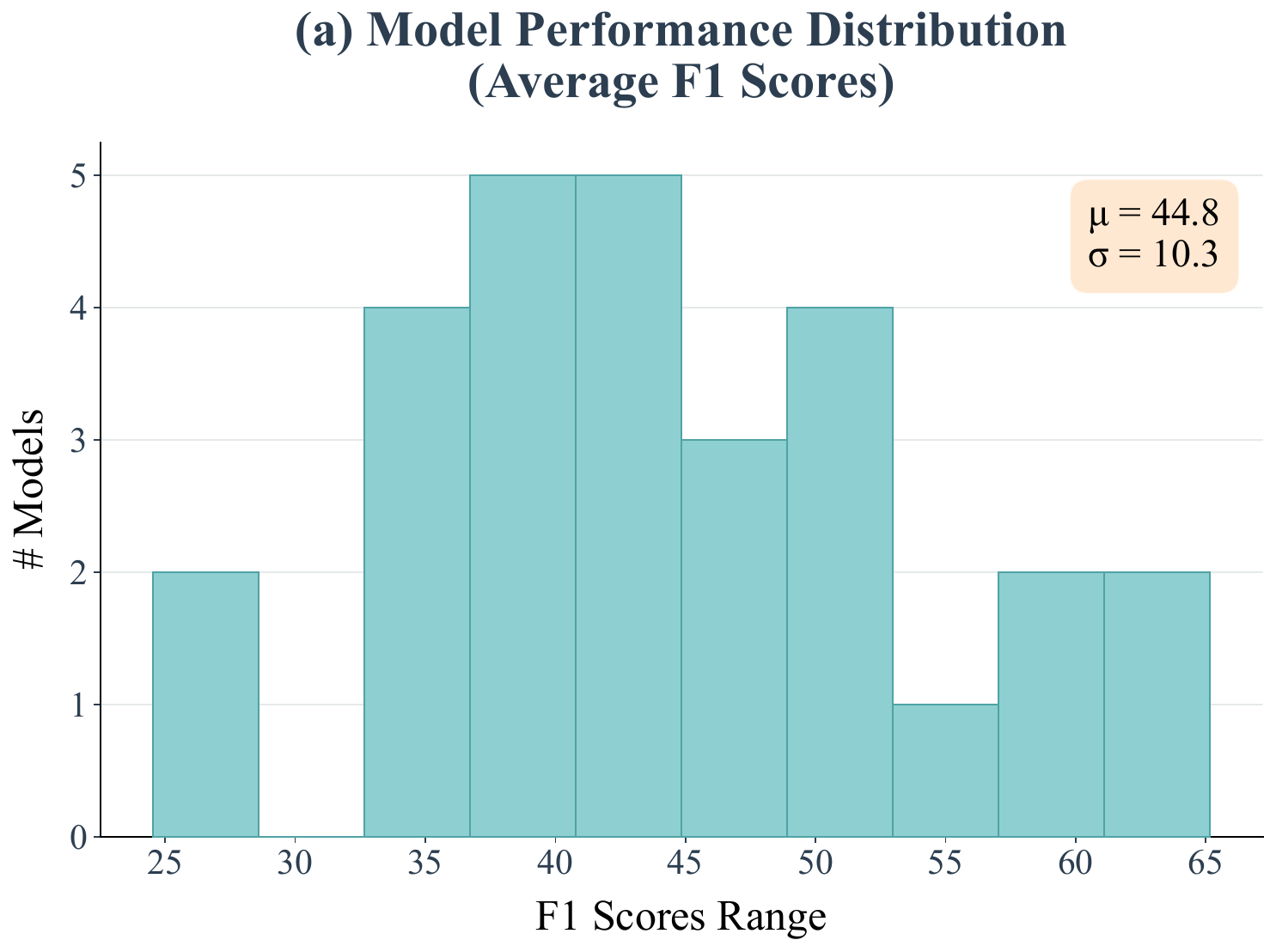}
    \end{subfigure}
    \begin{subfigure}[t]{0.48\linewidth}
        \includegraphics[width=\linewidth]{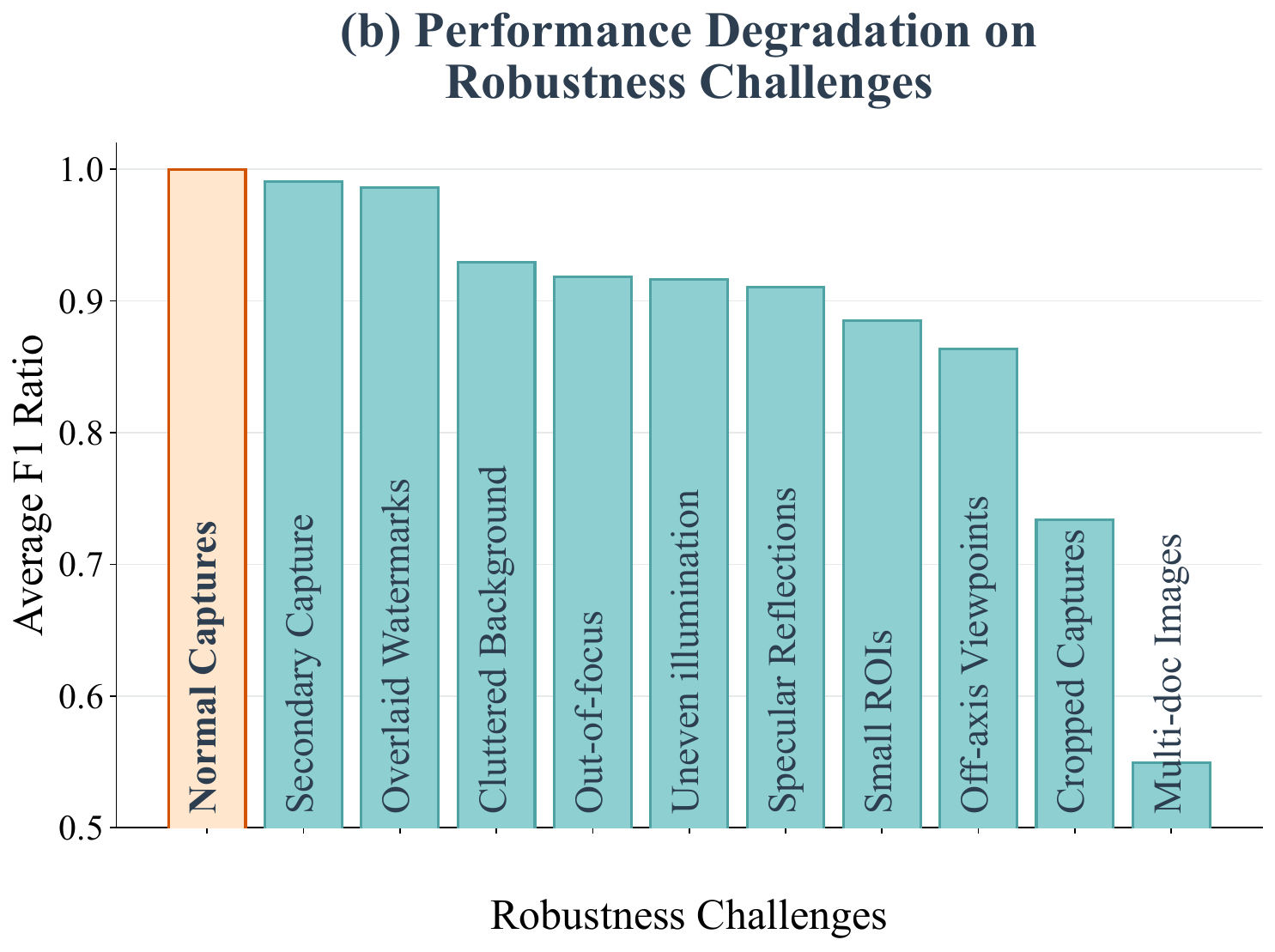}
    \end{subfigure}
    \vspace{1em}
    \begin{subfigure}[t]{0.48\linewidth}
        \includegraphics[width=\linewidth]{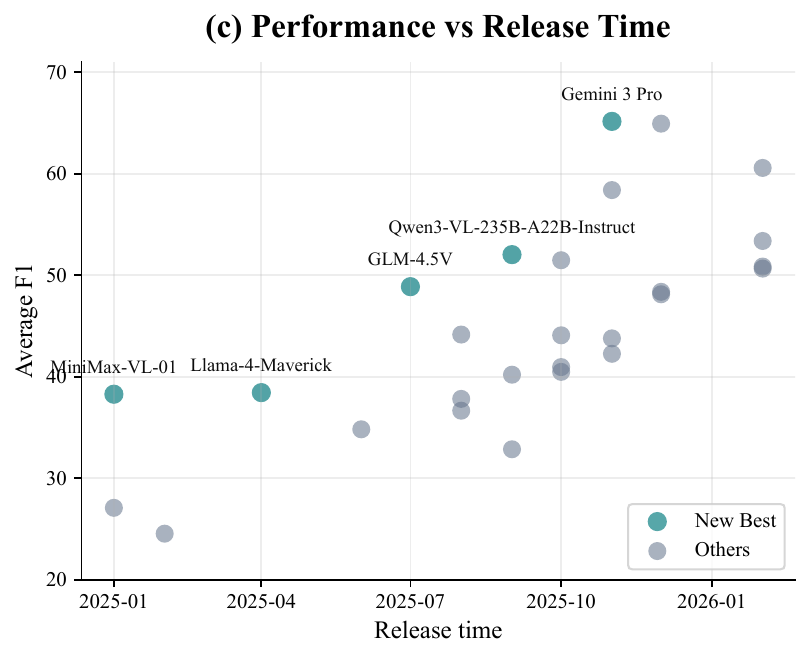}
    \end{subfigure}
    \begin{subfigure}[t]{0.48\linewidth}
        \includegraphics[width=\linewidth]{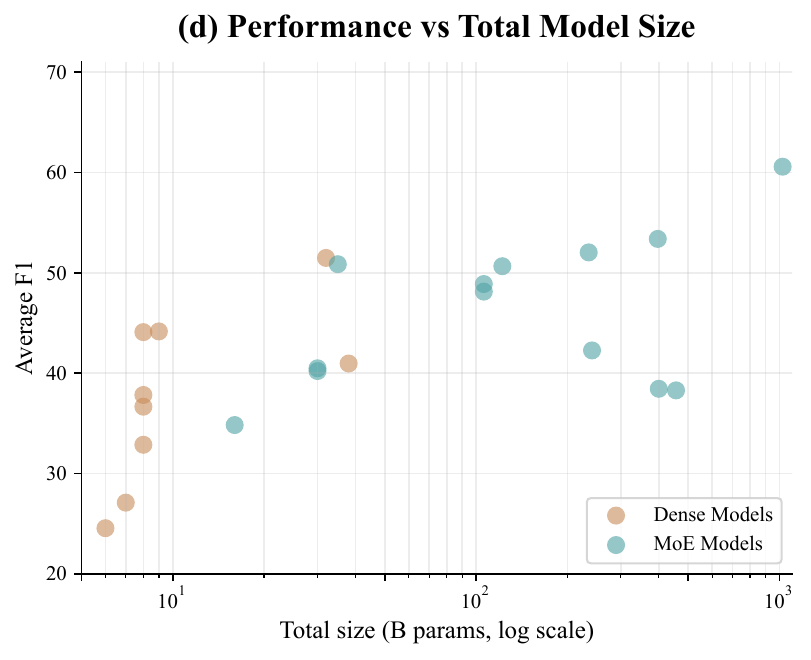}
    \end{subfigure}
    \caption{(a) The distribution of model performance has a wide spread; (b) Average model performance degrades when confronting the robustness challenges; (c) Best models have a considerable improvement over the year; (d) Scaling law still holds on this benchmark for both dense and MoE models.}
    \label{fig:performance_distribution_and_challenges}
\end{figure}

As shown in Figure~\ref{fig:performance_distribution_and_challenges}(a), model F1 scores span roughly 25–65, with a mean performance of $44.8\pm10.3$. The wide dispersion indicates that FCMBench is non-trivial and provides strong resolution for separating model capabilities. 

Figure~\ref{fig:performance_distribution_and_challenges}(b) further shows consistent performance drops across all models under robustness challenges relative to normal capture settings, highlighting the sensitivity of FCMBench to real-world robustness interference. Together, these results demonstrate that FCMBench offers a rigorous and reliable evaluation protocol for both overall competence and robustness of modern VLMs.

We observe a clear upward trend in performance over time (Figure~\ref{fig:performance_distribution_and_challenges}(c)). Since early 2025, the "best-in-class" performance has climbed from an F1 of $~40$ to over $60$, demonstrating rapid iterative improvements in VLMs. 

Figure~\ref{fig:performance_distribution_and_challenges}(d) confirms that scaling laws remain a reliable predictor of performance on our benchmark. Both Dense and Mixture-of-Experts (MoE) models show a positive correlation between total parameter count and average F1 score, with MoE models generally providing competitive performance at larger scales.

As shown in Table~\ref{tab:main_results}, Gemini 3 Pro and Gemini 3 Flash achieve the highest overall averages ($65.16$ and $64.94$, respectively). Among open-source models, Kimi-K2.5 and Qwen3.5-397B-A17B are the strongest baselines (overall averages are 60.58 and 53.38), which are comparable to other commercial models. The OCR-dependent pipeline (DeepSeek-OCR + V3.2) underperforms end-to-end VLMs in general (overall = 34.01), highlighting the fragility of two-stage approaches.

\begin{table}[!htb]
\newcommand{\best}[1]{\textbf{{#1}}} 
\newcommand{\second}[1]{\underline{#1}}
%\scriptsize
\small
\centering
\caption{Model performance by task. The best results in the tasks are indicated in \textbf{bold}, and the second-best results are marked with an \underline{underline}.}
\label{tab:main_results}
\centering
\begin{tabularx}{\linewidth}{l X X X X X X X c}
\toprule
& \multicolumn{3}{c}{\textbf{Perception}} 
& \multicolumn{4}{c}{\textbf{Reasoning}} 
& \multirow{2}{*}{\makecell[c]{\textbf{Overall}\\\textbf{Average}}}\\
\cmidrule(lr){2-4} \cmidrule(lr){5-8}
% 核心修改：IQE 移至 DTR 左侧
& \textbf{IQE} & \textbf{DTR} & \textbf{KIE} & \textbf{CC} & \textbf{VC} & \textbf{NC} & \textbf{RR} &  \\
\midrule
\multicolumn{3}{l}{\textbf{Commercial Models}} \\
\addlinespace[0.8em]
        
        Gemini 3 Pro & 44.83 & \second{90.68} & 43.36 & 60.68 & \best{87.27} & \best{59.75} & \best{69.56} & \best{65.16} \\
        
        Gemini 3 Flash & \best{52.19} & \best{95.32} & 43.99 & \best{69.93} & \second{73.06} & \second{57.26} & 62.82 & \second{64.94} \\
        
        Claude Opus 4.5 & 44.07 & 84.07 & 40.42 & 63.97 & 71.21 & 43.01 & 62.02 & 58.39 \\
        
        GPT 5.2 & 45.40 & 59.20 & 35.47 & 59.34 & 60.05 & 30.02 & 49.10 & 48.37 \\
        
        GPT 5.1 & 37.73 & 72.29 & 31.75 & 40.58 & 59.16 & 20.99 & 43.98 & 43.78 \\
        
\midrule
\multicolumn{3}{l}{\textbf{Open-Source Models}} \\
\addlinespace[0.8em]
        
        Kimi-K2.5 & 45.20 & 79.07 & \best{45.28} & 63.66 & 66.65 & 55.87 & \second{68.31} & 60.58 \\
        
        Qwen3.5-397B-A17B & 38.28 & 73.99 & 38.79 & 54.75 & 71.87 & 50.16 & 45.80 & 53.38 \\
        
        Qwen3-VL-235B-A22B-Instruct & 45.62 & 66.65 & 42.07 & 62.65 & 57.37 & 39.89 & 49.97 & 52.03 \\
        
        Qwen3-VL-32B-Instruct & \second{46.51} & 54.28 & 42.66 & \second{65.53} & 60.80 & 35.93 & 54.65 & 51.48 \\
        
        Qwen3.5-35B-A3B & 40.76 & 63.77 & 44.30 & 49.89 & 54.38 & 54.71 & 48.18 & 50.86 \\
        
        Qwen3.5-122B-A10B & 44.45 & 59.82 & \second{45.24} & 41.96 & 58.04 & 54.64 & 50.45 & 50.66 \\
        
        GLM-4.5V & 40.82 & 57.60 & 38.00 & 59.29 & 61.16 & 39.70 & 45.56 & 48.88 \\
        
        GLM-4.6V & 37.38 & 62.89 & 39.63 & 62.49 & 60.60 & 32.32 & 41.57 & 48.12 \\
        
        Ovis2.5 & 34.45 & 52.00 & 29.38 & 47.91 & 57.32 & 32.10 & 55.93 & 44.16 \\
        
        Qwen3-VL-8B-Instruct & 37.94 & 48.80 & 33.74 & 54.81 & 52.08 & 29.98 & 51.24 & 44.08 \\
        
        InternVL-3.5-241B-A28B & 31.93 & 61.27 & 32.84 & 43.90 & 54.46 & 26.48 & 44.94 & 42.26 \\
        
        InternVL-3.5-38B & 36.35 & 64.13 & 30.69 & 34.99 & 52.76 & 22.48 & 45.27 & 40.95 \\
        
        Qwen3-VL-30B-A3B-Instruct & 34.14 & 42.49 & 31.48 & 42.28 & 55.36 & 26.04 & 51.59 & 40.48 \\
        
        InternVL-3.5-30B-A3B & 34.73 & 50.87 & 28.31 & 39.47 & 55.03 & 21.84 & 51.18 & 40.20 \\
        
        Llama-4-Maverick & 41.43 & 37.47 & 27.50 & 34.68 & 59.78 & 27.64 & 40.54 & 38.44 \\
        
        MiniMax-VL-01 & 42.43 & 35.85 & 28.65 & 34.51 & 55.99 & 25.87 & 44.65 & 38.28 \\
        
        InternVL-3.5-8B & 26.44 & 50.69 & 26.72 & 40.63 & 55.86 & 19.46 & 44.94 & 37.82 \\
        
        MiniCPM-V 4.5 & 33.34 & 51.50 & 23.19 & 36.07 & 47.47 & 20.07 & 44.94 & 36.65 \\
        
        Kimi-VL & 21.65 & 47.16 & 26.59 & 35.36 & 52.49 & 15.56 & 44.94 & 34.82 \\
        DeepSeek(OCR+V3.2) & 10.57 & 43.05 & 29.00 & 29.51 & 49.48 & 23.73 & 52.76 & 34.01 \\
        LLaVA-OneVision-1.5 & 19.31 & 25.54 & 28.75 & 28.84 & 58.66 & 23.93 & 44.94 & 32.85 \\
        
        Janus-Pro & 11.77 & 10.78 & 10.87 & 56.84 & 39.67 & 14.73 & 44.94 & 27.09 \\
        
        Phi-4-multimodal-instruct & 14.44 & 24.46 & 20.84 & 28.11 & 37.23 & 0.87 & 45.80 & 24.54 \\          
\bottomrule
\end{tabularx}
\end{table}

\begin{figure}[t]
  \centering
   \includegraphics[width=0.9\linewidth]{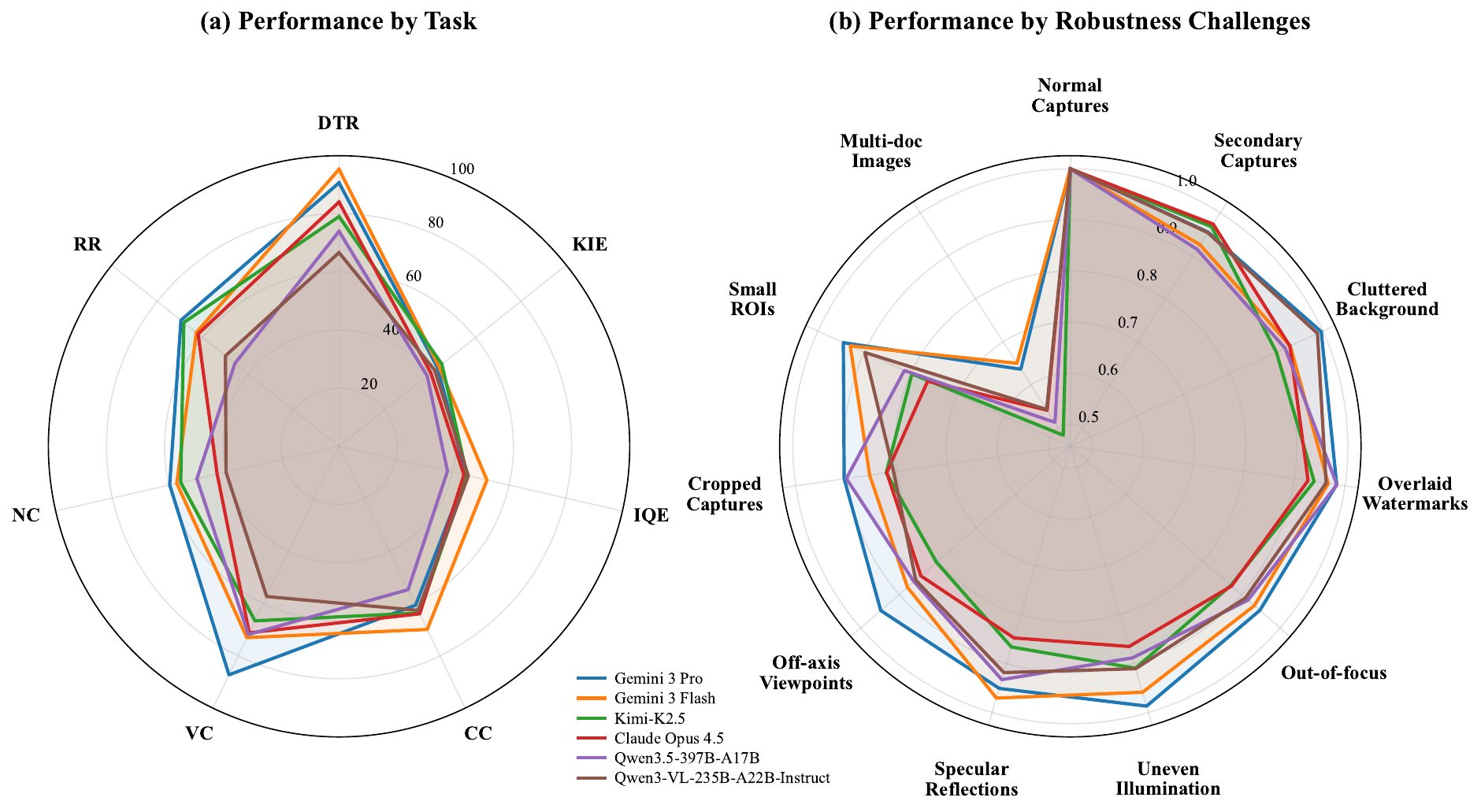}
  \caption{(a) Performance of best models by tasks. (b) Performance degradation of best models on different challenges. Scores are scaled so that each model's scores on Normal Captures are 1.0.}
  \label{fig:two-side-by-side}
\end{figure}

\subsubsection{Task-Specific Performance}
Model performance varies significantly when partitioned between Perception (IQE, DTR, KIE) and Reasoning (CC, VC, NC, RR) tasks. Most top-tier models excel in DTR, with Gemini 3 Flash reaching a peak score of $95.32$ (Table 1). However, performance drops sharply on KIE and IQE tasks, where even the best models struggle to exceed $50$. 

The radar plot in Figure~\ref{fig:two-side-by-side}(a) highlights a clear contrast between perception-oriented extraction and credit-oriented reasoning on FCMBench. For the strongest models, KIE performance is relatively clustered (e.g., Gemini~3~Pro/Flash at 43.36/43.99 and Kimi-K2.5 at 45.28), suggesting that state-of-the-art VLMs have converged to a comparable level of key information extraction under our templates. In contrast, the reasoning dimensions exhibit substantially larger variance. In other words, even when models can “read” similar key fields at limited image quality, they differ in whether they can reliably integrate visual evidence, apply domain constraints, and remain robust under real-world interference. 

These observations indicate that (i) FCMBench reasoning tasks are not a trivial extension of extraction accuracy, and (ii) the benchmark provides meaningful separation among SOTA models precisely in the decision-relevant stages of credit assessment. 

\subsubsection{Performance over Robustness Challenges}
We are concerned that some images may suffer from severe acquisition artifacts (e.g., out-of-focus blur and specular reflection), which could render required fields partially unreadable even for human annotators. Since the ground truths were pre-generated without accounting for such artifacts, absolute accuracy in these cases might be limited by image quality rather than model capability. For this reason, to ensure fair robustness comparison, we report each model’s performance relative to “Normal Captures” in Figure~\ref{fig:performance_distribution_and_challenges}(b). We further report the same relative performance metrics of top commercial models and open-source models in Figure~\ref{fig:two-side-by-side} (b). This gap between the clustered low KIE scores and widely spread reasoning performance further indicates that FCMBench robustness interferences are both challenging and discriminative.. 

Both of these views suggest that high-performing models are not inherently robust. The radar plot shows that even the best models remain vulnerable to specific robustness interference, and their margins can shrink substantially under the hardest conditions. In practice, this means that a model that performs strongly overall may still fail systematically for certain acquisition patterns (\eg, partial crops, multi-doc compositions, or severe blur), which are common in user uploads and field collection.

These findings motivate robustness as a first-class pre-deployment requirement for business settings. Before integrating VLMs into credit workflows, developers should explicitly evaluate robustness on representative capture noise and enforce mitigation strategies such as capture-side guidelines (re-capture prompts, multi-view collection), automated quality gates (blur/ROI/crop/multi-doc detection), and targeted data augmentation or finetuning on the failure modes that dominate degradation.

\section{Conclusion and Future Work}
This work presents FCMBench, the first large-scale multimodal benchmark dedicated to financial credit-risk workflows. We construct a set of highly realistic certificate images and build FCMBench, which comprises 26 certificate types, 5,198 images, 13,806 QA pairs, and 7 core application-oriented tasks aligned with the credit review process. By open-sourcing the dataset and evaluation code, FCMBench fills the gap in domain-specific multimodal benchmarks for the credit field and enables effective evaluation of perception and reasoning capabilities under real-world robustness challenges.

However, FCMBench still has broad improvement space. In terms of data coverage, it only involves physical certificate photos, lacking common formats such as screenshots, scanned copies and digital certificates, while its certificate and template categories need further expansion; introducing non-image modalities including audio and video is also a meaningful direction. %For task and robustness design, the benchmark only covers core credit-review workflows, leaving tasks like image grounding and tampering detection unexplored, and does not include real-scenario challenges such as handwritten documents.

FCMBench will be continuously updated and expanded to better match the complexity and practical demands of modern credit-risk workflows. We hope this benchmark can promote collaborative innovation between academia and industry, and accelerate the research and development of reliable and practical credit AI systems.

%\clearpage  % TODO FINAL: This \clearpage needs to be removed from both review and camera-ready versions.

\section*{Acknowledgements}
The authors would like to thank Chaoyou Fu from Nanjing University, Didi Hu, Yaoxuan Wang, Huifang Du, Mengyuan Liu, Runze Cui, Mengjie Li, and other colleagues at Qfin Tech Inc. for their valuable assistance and insightful inspiration during the development of this benchmark.

\bibliographystyle{unsrt}
\bibliography{main}

@article{chen2024mmstar,
    title={Are We on the Right Way for Evaluating Large Vision-Language Models?},
    author={Chen, Lin and Li, Jinsong and Dong, Xiaoyi and Zhang, Pan and Zang, Yuhang and Chen, Zehui and Duan, Haodong and Wang, Jiaqi and Qiao, Yu and Lin, Dahua and others},
    journal={arXiv preprint arXiv:2403.20330},
    year={2024}
  }

@misc{fu2025mmecomprehensiveevaluationbenchmark,
      title={MME: A Comprehensive Evaluation Benchmark for Multimodal Large Language Models}, 
      author={Chaoyou Fu and Peixian Chen and Yunhang Shen and Yulei Qin and Mengdan Zhang and Xu Lin and Jinrui Yang and Xiawu Zheng and Ke Li and Xing Sun and Yunsheng Wu and Rongrong Ji and Caifeng Shan and Ran He},
      year={2025},
      eprint={2306.13394},
      archivePrefix={arXiv},
      primaryClass={cs.CV},
      url={https://arxiv.org/abs/2306.13394}, 
}

@misc{zhu2024mmdocbenchbenchmark,
      title={MMDocBench: Benchmarking Large Vision-Language Models for Fine-Grained Visual Document Understanding}, 
      author={Fengbin Zhu and Ziyang Liu and Xiang Yao Ng and Haohui Wu and Wenjie Wang and Fuli Feng and Chao Wang and Huanbo Luan and Tat Seng Chua},
      year={2024},
      eprint={2410.21311},
      archivePrefix={arXiv},
      primaryClass={cs.CV},
      url={https://arxiv.org/abs/2410.21311}, 
}

@misc{tang2025chartmuseum,
      title={ChartMuseum: Testing Visual Reasoning Capabilities of Large Vision-Language Models}, 
      author={Liyan Tang and Grace Kim and Xinyu Zhao and Thom Lake and Wenxuan Ding and Fangcong Yin and Prasann Singhal and Manya Wadhwa and Zeyu Leo Liu and Zayne Sprague and Ramya Namuduri and Bodun Hu and Juan Diego Rodriguez and Puyuan Peng and Greg Durrett},
      year={2025},
      eprint={2505.13444},
      archivePrefix={arXiv},
      primaryClass={cs.CL},
      url={https://arxiv.org/abs/2505.13444}, 
}

@misc{yue2024MMMU,
      title={MMMU: A Massive Multi-discipline Multimodal Understanding and Reasoning Benchmark for Expert AGI}, 
      author={Xiang Yue and Yuansheng Ni and Kai Zhang and Tianyu Zheng and Ruoqi Liu and Ge Zhang and Samuel Stevens and Dongfu Jiang and Weiming Ren and Yuxuan Sun and Cong Wei and Botao Yu and Ruibin Yuan and Renliang Sun and Ming Yin and Boyuan Zheng and Zhenzhu Yang and Yibo Liu and Wenhao Huang and Huan Sun and Yu Su and Wenhu Chen},
      year={2024},
      eprint={2311.16502},
      archivePrefix={arXiv},
      primaryClass={cs.CL},
      url={https://arxiv.org/abs/2311.16502}, 
}

@misc{yue2025mmmupro,
      title={MMMU-Pro: A More Robust Multi-discipline Multimodal Understanding Benchmark},
      author={Xiang Yue and Tianyu Zheng and Yuansheng Ni and Yubo Wang and Kai Zhang and Shengbang Tong and Yuxuan Sun and Botao Yu and Ge Zhang and Huan Sun and Yu Su and Wenhu Chen and Graham Neubig},
      year={2025},
      eprint={2409.02813},
      archivePrefix={arXiv},
      primaryClass={cs.CL},
      url={https://arxiv.org/abs/2409.02813}, 
}

@article{Liu2024OCRBench,
   title={OCRBench: on the hidden mystery of OCR in large multimodal models},
   volume={67},
   ISSN={1869-1919},
   url={http://dx.doi.org/10.1007/s11432-024-4235-6},
   DOI={10.1007/s11432-024-4235-6},
   number={12},
   journal={Science China Information Sciences},
   publisher={Springer Science and Business Media LLC},
   author={Liu, Yuliang and Li, Zhang and Huang, Mingxin and Yang, Biao and Yu, Wenwen and Li, Chunyuan and Yin, Xu-Cheng and Liu, Cheng-Lin and Jin, Lianwen and Bai, Xiang},
   year={2024},
   month=dec }

@misc{wang2025wilddoc,
      title={WildDoc: How Far Are We from Achieving Comprehensive and Robust Document Understanding in the Wild?}, 
      author={An-Lan Wang and Jingqun Tang and Liao Lei and Hao Feng and Qi Liu and Xiang Fei and Jinghui Lu and Han Wang and Weiwei Liu and Hao Liu and Yuliang Liu and Xiang Bai and Can Huang},
      year={2025},
      eprint={2505.11015},
      archivePrefix={arXiv},
      primaryClass={cs.CV},
      url={https://arxiv.org/abs/2505.11015}, 
}

@misc{masry2025chartqaprodiversechallengingbenchmark,
      title={ChartQAPro: A More Diverse and Challenging Benchmark for Chart Question Answering}, 
      author={Ahmed Masry and Mohammed Saidul Islam and Mahir Ahmed and Aayush Bajaj and Firoz Kabir and Aaryaman Kartha and Md Tahmid Rahman Laskar and Mizanur Rahman and Shadikur Rahman and Mehrad Shahmohammadi and Megh Thakkar and Md Rizwan Parvez and Enamul Hoque and Shafiq Joty},
      year={2025},
      eprint={2504.05506},
      archivePrefix={arXiv},
      primaryClass={cs.CL},
      url={https://arxiv.org/abs/2504.05506}, 
}

@inproceedings{xie2024idnet,
  title={IDNet: A Novel Identity Document Dataset via Few-Shot and Quality-Driven Synthetic Data Generation},
  author={Xie, Lulu and Wang, Yancheng and Guan, Hong and Nag, Soham and Goel, Rajeev and Swamy, Niranjan and Yang, Yingzhen and Xiao, Chaowei and Prisby, Jonathan and Maciejewski, Ross and others},
  booktitle={2024 IEEE International Conference on Big Data (BigData)},
  pages={2244--2253},
  year={2024},
  organization={IEEE}
}

@misc{msswift2024,
      title={SWIFT:A Scalable lightWeight Infrastructure for Fine-Tuning},
      author={Yuze Zhao and Jintao Huang and Jinghan Hu and Xingjun Wang and Yunlin Mao and Daoze Zhang and Zeyinzi Jiang and Zhikai Wu and Baole Ai and Ang Wang and Wenmeng Zhou and Yingda Chen},
      year={2024},
      eprint={2408.05517},
      archivePrefix={arXiv},
      primaryClass={cs.CL},
      url={https://arxiv.org/abs/2408.05517},
}

@misc{li2025cfbenchmarkmmchinesefinancialassistant,
      title={CFBenchmark-MM: Chinese Financial Assistant Benchmark for Multimodal Large Language Model},
      author={Jiangtong Li and Yiyun Zhu and Dawei Cheng and Zhijun Ding and Changjun Jiang},
      year={2025},
      eprint={2506.13055},
      archivePrefix={arXiv},
      primaryClass={cs.CL},
      url={https://arxiv.org/abs/2506.13055}, 
}

@misc{yanglet2025multimodalfinancialfoundationmodels,
      title={Multimodal Financial Foundation Models (MFFMs): Progress, Prospects, and Challenges},
      author={Xiao-Yang Liu Yanglet and Yupeng Cao and Li Deng},
      year={2025},
      eprint={2506.01973},
      archivePrefix={arXiv},
      primaryClass={cs.CE},
      url={https://arxiv.org/abs/2506.01973}, 
}

@misc{liu2025visfinevalscenariodrivenchinesemultimodal,
      title={VisFinEval: A Scenario-Driven Chinese Multimodal Benchmark for Holistic Financial Understanding},
      author={Zhaowei Liu and Xin Guo and Haotian Xia and Lingfeng Zeng and Fangqi Lou and Jinyi Niu and Mengping Li and Qi Qi and Jiahuan Li and Wei Zhang and Yinglong Wang and Weige Cai and Weining Shen and Liwen Zhang},
      year={2025},
      eprint={2508.09641},
      archivePrefix={arXiv},
      primaryClass={cs.CE},
      url={https://arxiv.org/abs/2508.09641}, 
}

@misc{luo2025finmmebenchmarkdatasetfinancial,
      title={FinMME: Benchmark Dataset for Financial Multi-Modal Reasoning Evaluation},
      author={Junyu Luo and Zhizhuo Kou and Liming Yang and Xiao Luo and Jinsheng Huang and Zhiping Xiao and Jingshu Peng and Chengzhong Liu and Jiaming Ji and Xuanzhe Liu and Sirui Han and Ming Zhang and Yike Guo},
      year={2025},
      eprint={2505.24714},
      archivePrefix={arXiv},
      primaryClass={cs.CL},
      url={https://arxiv.org/abs/2505.24714}, 
}

@misc{peng2025multifinbenbenchmarkinglargelanguage,
      title={MultiFinBen: Benchmarking Large Language Models for Multilingual and Multimodal Financial Application},
      author={Xueqing Peng and Lingfei Qian and Yan Wang and Ruoyu Xiang and Yueru He and Yang Ren and Mingyang Jiang and Vincent Jim Zhang and Yuqing Guo and Jeff Zhao and Huan He and Yi Han and Yun Feng and Yuechen Jiang and Yupeng Cao and Haohang Li and Yangyang Yu and Xiaoyu Wang and Penglei Gao and Shengyuan Lin and Keyi Wang and Shanshan Yang and Yilun Zhao and Zhiwei Liu and Peng Lu and Jerry Huang and Suyuchen Wang and Triantafillos Papadopoulos and Polydoros Giannouris and Efstathia Soufleri and Nuo Chen and Zhiyang Deng and Heming Fu and Yijia Zhao and Mingquan Lin and Meikang Qiu and Kaleb E Smith and Arman Cohan and Xiao-Yang Liu and Jimin Huang and Guojun Xiong and Alejandro Lopez-Lira and Xi Chen and Junichi Tsujii and Jian-Yun Nie and Sophia Ananiadou and Qianqian Xie},
      year={2025},
      eprint={2506.14028},
      archivePrefix={arXiv},
      primaryClass={cs.CL},
      url={https://arxiv.org/abs/2506.14028}, 
}

@misc{deng2025finmrknowledgeintensivemultimodalbenchmark,
      title={FinMR: A Knowledge-Intensive Multimodal Benchmark for Advanced Financial Reasoning},
      author={Shuangyan Deng and Haizhou Peng and Jiachen Xu and Rui Mao and Ciprian Doru Giurcăneanu and Jiamou Liu},
      year={2025},
      eprint={2510.07852},
      archivePrefix={arXiv},
      primaryClass={cs.AI},
      url={https://arxiv.org/abs/2510.07852}, 
}

@misc{gpt5_1,
  title = {GPT-5.1},
  author = {OpenAI},
  year = {2025},
  howpublished = {\url{https://openai.com/index/gpt-5-1/}},
  note = {Accessed: 2025-11-12}
}

@misc{qwen3.5,
  url = {https://github.com/QwenLM/Qwen3.5}
}

@misc{gemini3,
  title = {Gemini 3 Pro},
  author = {Google},
  year = {2025},
  howpublished = {\url{https://aistudio.google.com/models/gemini-3}},
  note = {Accessed: 2025-11-18}
}

@misc{claude_opus,
  title = {Claude-Opus-4.5},
  author = {Anthropic},
  year = {2025},
  howpublished = {\url{https://www.anthropic.com/news/claude-opus-4-5}},
  note = {Accessed: 2025-11-24}
}

@article{qwen3vl,
      title={Qwen3-VL Technical Report}, 
      author={Shuai Bai and Yuxuan Cai and Ruizhe Chen and Keqin Chen and Xionghui Chen and Zesen Cheng and Lianghao Deng and Wei Ding and Chang Gao and Chunjiang Ge and others},
      journal={arXiv preprint arXiv:2511.21631},
      year={2025}
}

@article{internvl35,
  title={InternVL3.5: Advancing Open-Source Multimodal Models in Versatility, Reasoning, and Efficiency},
  author={Wang, Weiyun and Gao, Zhangwei and Gu, Lixin and Pu, Hengjun and Cui, Long and Wei, Xingguang and Liu, Zhaoyang and Jing, Linglin and Ye, Shenglong and others},
  journal={arXiv preprint arXiv:2508.18265},
  year={2025}
}

@misc{glm45v,
      title={GLM-4.5V and GLM-4.1V-Thinking}, 
      author={V Team and Wenyi Hong and Wenmeng Yu and Xiaotao Gu and others},
      year={2025},
      eprint={2507.01006},
      archivePrefix={arXiv},
      primaryClass={cs.CV}
}

@article{deepseekocr,
  title={DeepSeek-OCR: Contexts Optical Compression},
  author={Wei, Haoran and Sun, Yaofeng and Li, Yukun},
  journal={arXiv preprint arXiv:2510.18234},
  year={2025}
}

@misc{januspro,
      title={Janus-Pro: Unified Multimodal Understanding and Generation}, 
      author={Xiaokang Chen and Zhiyu Wu and Xingchao Liu and Zizheng Pan and Wen Liu and Zhenda Xie and Xingkai Yu and Chong Ruan},
      year={2025},
      eprint={2501.17811},
      archivePrefix={arXiv}
}

@misc{minimaxvl,
      title={MiniMax-01: Scaling Foundation Models with Lightning Attention}, 
      author={MiniMax and Aonian Li and Bangwei Gong and Bo Yang and Boji Shan and others},
      year={2025},
      eprint={2501.08313},
      archivePrefix={arXiv}
}

@misc{phi4mm,
      title={Phi-4-Mini Technical Report: Compact yet Powerful Multimodal Language Models}, 
      author={Microsoft and Abdelrahman Abouelenin and Atabak Ashfaq and Adam Atkinson and others},
      year={2025},
      eprint={2503.01743},
      archivePrefix={arXiv}
}

@misc{kimivl,
      title={Kimi-VL Technical Report}, 
      author={Kimi Team et al.},
      year={2025},
      eprint={2504.07491},
      archivePrefix={arXiv}
}

@misc{kimik2.5,
      title={Kimi K2.5: Visual Agentic Intelligence}, 
      author={Kimi Team et al.},
      year={2026},
      eprint={2602.02276},
      archivePrefix={arXiv},
      primaryClass={cs.CL},
      url={https://arxiv.org/abs/2602.02276}, 
}

@misc{minicpmv45,
      title={MiniCPM-V 4.5: Cooking Efficient MLLMs via Architecture, Data, and Training Recipe}, 
      author={Tianyu Yu and Zefan Wang and Chongyi Wang and Fuwei Huang and others},
      year={2025},
      eprint={2509.18154},
      archivePrefix={arXiv}
}

@misc{ovis25,
      title={Ovis2.5 Technical Report}, 
      author={Shiyin Lu and Yang Li and Yu Xia and Yuwei Hu and Shanshan Zhao and Yanqing Ma and Zhichao Wei and Yinglun Li and others},
      year={2025},
      eprint={2508.11737},
      archivePrefix={arXiv}
}

@inproceedings{llavaonevision,
  title={LLaVA-OneVision-1.5: Fully Open Framework for Democratized Multimodal Training},
  author={An, Xiang and Xie, Yin and Yang, Kaicheng and Zhang, Wenkang and Zhao, Xiuwei and Cheng, Zheng and others},
  booktitle={arXiv},  
  year={2025}
 }

@misc{deepseek3.2,
      title={DeepSeek-V3.2: Pushing the Frontier of Open Large Language Models}, 
      author={DeepSeek-AI et al.},
      year={2025},
      eprint={2512.02556},
      archivePrefix={arXiv},
      primaryClass={cs.CL},
      url={https://arxiv.org/abs/2512.02556}, 
}

@misc{llama4,
  title = {The Llama 4 herd},
  author = {Meta AI},
  year = {2025},
  howpublished = {\url{https://ai.meta.com/blog/llama-4-multimodal-intelligence/}},
  note = {Accessed: 2025-04-05}
}
\clearpage
\appendix
\renewcommand{\thefigure}{A\arabic{figure}}
\renewcommand{\thetable}{A\arabic{table}}
\section{Appendix}
\subsection{Task Mapping and Case Examples}
See the web page \href{https://qfin-tech.github.io/FCMBench/Examples.html}{Examples.html} for how the FCMBench task hierarchy relates to the real-world credit review workflow. Hover over the workflow, and the related tasks will be highlighted, and vice versa. Click a task to view its subtasks, and a subtask to access its case examples (prompt-image-answer).

\subsection{Results on Different Languages}
% Results table Chinese
Table~\ref{tab:chinese_results} and Table~\ref{tab:english_results} shows the model performance separately on the Chinese and English prompts/certificates. Top Chinese open-source models show reduced gap vs. commercial models in Chinese prompts compared with in English.

\begin{table}[!htb]
\newcommand{\best}[1]{\textbf{{#1}}} 
\newcommand{\second}[1]{\underline{#1}}
\scriptsize
\centering
\caption{Model performance by task on \textbf{Chinese} prompts and documents. The best results in the tasks are indicated in \textbf{bold}, and the second-best results are marked with an \underline{underline}.}
\label{tab:chinese_results}
\centering
\begin{tabularx}{\linewidth}{l XXXXXXX c}
\toprule
& \multicolumn{3}{c}{\textbf{Perception}}
& \multicolumn{4}{c}{\textbf{Reasoning}} 
& \multirow{2}{*}{\makecell[c]{\textbf{Overall}\\\textbf{Average}}}\\
\cmidrule(lr){2-4} \cmidrule(lr){5-8}
% 核心修改：IQE 移至 DTR 左侧
& \textbf{IQE} & \textbf{DTR} & \textbf{KIE} & \textbf{CC} & \textbf{VC} & \textbf{NC} & \textbf{RR} &  \\
\midrule
\multicolumn{3}{l}{\textbf{Commercial Models}} \\
\addlinespace[0.8em]
        
        Gemini 3 Flash & \best{48.18} & \best{93.24} & 30.30 & \best{80.55} & 81.62 & \second{42.39} & 62.82 & \best{65.15} \\
        
        Gemini 3 Pro & \second{46.68} & 88.58 & 30.53 & 65.78 & \best{89.80} & \best{46.05} & \best{69.56} & \second{65.05} \\
        
        Claude Opus 4.5 & 41.04 & 87.82 & 28.95 & 73.41 & \second{84.35} & 27.70 & 62.02 & 60.71 \\
        
        GPT 5.2 & 43.68 & 64.50 & 25.81 & 71.15 & 71.19 & 18.90 & 49.10 & 50.11 \\
        GPT 5.1 & 38.54 & 67.69 & 27.27 & 47.42 & 69.64 & 8.90 & 43.98 & 44.15 \\
        
\midrule
\multicolumn{3}{l}{\textbf{Open-Source Models}} \\
\addlinespace[0.8em]
        
        Kimi-K2.5 & 44.11 & 90.07 & 33.10 & 75.93 & 78.83 & 41.86 & \second{68.31} & 64.68 \\

        Qwen3-VL-235B-A22B-Instruct & 44.97 & \second{92.36} & 32.15 & 70.84 & 59.33 & 23.66 & 49.97 & 54.72 \\
        
        Qwen3-VL-32B-Instruct & 44.54 & 64.98 & 30.01 & \second{77.29} & 72.03 & 23.51 & 54.65 & 53.75 \\
        
        Qwen3.5-397B-A17B & 34.69 & 71.00 & 27.27 & 58.62 & 72.60 & 36.33 & 45.80 & 51.94 \\
        
        Qwen3.5-122B-A10B & 42.61 & 68.34 & \second{34.03} & 41.59 & 65.63 & 42.21 & 50.45 & 50.38 \\

        GLM-4.5V & 34.26 & 62.13 & 27.03 & 68.58 & 74.29 & 29.95 & 45.56 & 51.26 \\
        
        Ovis2.5 & 40.26 & 75.87 & 26.47 & 59.95 & 62.97 & 19.04 & 55.93 & 50.04 \\
        
        Qwen3.5-35B-A3B & 34.69 & 62.77 & \best{35.52} & 56.70 & 57.59 & 41.94 & 48.18 & 50.45 \\
        
        GLM-4.6V & 30.41 & 75.21 & 28.03 & 72.30 & 72.59 & 15.79 & 41.57 & 50.92 \\
        
        Qwen3-VL-8B-Instruct & 38.97 & 55.78 & 27.45 & 61.69 & 52.84 & 17.08 & 51.24 & 44.35 \\

        InternVL-3.5-241B-A28B & 29.98 & 72.06 & 23.13 & 53.70 & 65.66 & 13.16 & 44.94 & 45.44 \\
        
        InternVL-3.5-30B-A3B & 32.55 & 62.91 & 21.00 & 50.98 & 67.23 & 6.60 & 51.18 & 43.31 \\
        
        InternVL-3.5-8B & 28.91 & 62.66 & 21.64 & 55.58 & 68.76 & 6.74 & 44.94 & 43.38 \\
        
        MiniMax-VL-01 & 43.25 & 47.65 & 20.09 & 47.22 & 70.57 & 13.72 & 44.65 & 40.65 \\
        
        InternVL-3.5-38B & 32.76 & 76.03 & 21.43 & 41.35 & 59.49 & 8.42 & 45.27 & 42.00 \\
        
        Qwen3-VL-30B-A3B-Instruct & 28.05 & 50.99 & 23.19 & 45.25 & 63.42 & 11.05 & 51.59 & 40.91 \\
        
        MiniCPM-V 4.5 & 31.69 & 68.53 & 16.35 & 47.28 & 55.65 & 5.34 & 44.94 & 39.68 \\
        
        Kimi-VL & 24.84 & 54.89 & 25.04 & 50.46 & 61.01 & 0.83 & 44.94 & 39.53 \\
        
        Llama-4-Maverick & 39.61 & 44.50 & 18.24 & 36.67 & 69.72 & 12.15 & 40.54 & 36.97 \\
        
        LLaVA-OneVision-1.5 & 26.77 & 35.41 & 25.63 & 38.60 & 50.97 & 14.09 & 44.94 & 34.94 \\
        
        DeepSeek-OCR + DeepSeek V3.2 & 9.85 & 43.46 & 24.23 & 32.59 & 47.34 & 11.61 & 52.76 & 35.33 \\
        
        Janus-Pro & 11.13 & 11.53 & 8.97 & 59.32 & 44.84 & 0.78 & 44.94 & 28.40 \\
        
        Phi-4-multimodal-instruct & 8.78 & 21.07 & 17.89 & 33.84 & 36.90 & 1.13 & 45.80 & 26.10 \\      
\bottomrule
\end{tabularx}
\end{table}

% Results table English
\begin{table}[!htb]
\newcommand{\best}[1]{\textbf{{#1}}} 
\newcommand{\second}[1]{\underline{#1}}
\scriptsize
\centering
\caption{Model performance by task on \textbf{English} prompts and documents. The best results in the tasks are indicated in \textbf{bold}, and the second-best results are marked with an \underline{underline}.}
\label{tab:english_results}
\centering
\begin{tabularx}{\linewidth}{l X X X X X X c}
\toprule
& \multicolumn{3}{c}{\textbf{Perception}} 
& \multicolumn{3}{c}{\textbf{Reasoning}} 
& \multirow{2}{*}{\makecell[c]{\textbf{Overall}\\\textbf{Average}}}\\
\cmidrule(lr){2-4} \cmidrule(lr){5-7}
% 核心修改：IQE 移至 DTR 左侧
& \textbf{IQE} & \textbf{DTR} & \textbf{KIE} & \textbf{CC} & \textbf{VC} & \textbf{NC} &  \\
\midrule
\multicolumn{3}{l}{\textbf{Commercial Models}} \\
\addlinespace[0.8em]
        
        Gemini 3 Pro & 42.98 & \second{92.78} & 59.41 & \second{50.50} & \best{84.74} & \best{78.02} & \best{73.09} \\
        
        Gemini 3 Flash & \best{56.20} & \best{97.41} & \best{61.10} & 48.67 & 64.50 & \second{77.07} & \second{69.75} \\
        
        Claude Opus 4.5 & 47.11 & 80.31 & 54.76 & 45.09 & 58.07 & 63.42 & 60.33 \\
        
        GPT 5.2 & 47.11 & 53.90 & 47.53 & 35.74 & 48.91 & 44.84 & 46.18 \\
        
        GPT 5.1 & 36.91 & 76.88 & 37.36 & 26.88 & 48.68 & 37.10 & 45.38 \\
        
\midrule
\multicolumn{3}{l}{\textbf{Open-Source Models}} \\
\addlinespace[0.8em]
        
        Qwen3.5-397B-A17B & 41.87 & 76.98 & 53.18 & 47.01 & \second{71.15} & 68.59 & 63.38 \\

        Kimi-K2.5 & 46.28 & 68.07 & \second{60.51} & 39.11 & 54.48 & 74.54 & 59.34 \\
        
        Qwen3.5-35B-A3B & 46.83 & 64.76 & 55.27 & 36.27 & 51.17 & 71.73 & 55.84 \\
        
        Qwen3.5-122B-A10B & 46.28 & 51.30 & 59.25 & 42.71 & 50.45 & 71.20 & 54.98 \\
        
        Qwen3-VL-235B-A22B-Instruct & 46.28 & 40.95 & 54.47 & 46.28 & 55.42 & 61.52 & 51.73 \\
        
        GLM-4.6V & 44.35 & 50.56 & 54.13 & 42.89 & 48.61 & 54.35 & 50.11 \\
        
        Qwen3-VL-32B-Instruct & \second{48.48} & 43.57 & 58.47 & 42.01 & 49.57 & 52.47 & 49.22 \\
        
        GLM-4.5V & 47.38 & 53.08 & 51.71 & 40.71 & 48.04 & 52.69 & 49.25 \\

        Qwen3-VL-8B-Instruct & 36.91 & 41.83 & 41.60 & 41.03 & 51.31 & 47.18 & 44.59 \\
        
        Qwen3-VL-30B-A3B-Instruct & 40.22 & 33.98 & 41.83 & 36.34 & 47.29 & 46.03 & 41.10 \\
        
        InternVL-3.5-38B & 39.94 & 52.24 & 42.26 & 22.26 & 46.03 & 41.22 & 40.80 \\
        
        Llama-4-Maverick & 43.25 & 30.45 & 39.08 & 30.71 & 49.84 & 48.30 & 39.67 \\
        
        InternVL-3.5-241B-A28B & 33.88 & 50.49 & 44.97 & 24.30 & 43.26 & 44.24 & 41.45 \\
        
        Ovis2.5 & 28.65 & 28.13 & 33.02 & 23.84 & 51.67 & 49.51 & 37.23 \\
        
        InternVL-3.5-30B-A3B & 36.91 & 38.83 & 37.45 & 16.45 & 42.82 & 42.15 & 35.54 \\
        
        DeepSeek-OCR + DeepSeek V3.2 & 11.29 & 42.64 & 34.97 & 23.33 & 51.61 & 39.88 & 38.49 \\
        
        MiniMax-VL-01 & 41.60 & 24.05 & 39.34 & 9.09 & 41.41 & 42.06 & 31.19 \\
        
        MiniCPM-V 4.5 & 34.99 & 34.47 & 31.74 & 13.66 & 39.28 & 39.69 & 31.77 \\
        
        InternVL-3.5-8B & 23.97 & 38.72 & 33.07 & 10.72 & 42.96 & 36.43 & 32.38 \\
        
        LLaVA-OneVision-1.5 & 11.85 & 15.67 & 32.65 & 9.33 & 66.34 & 37.05 & 32.21 \\
        
        Kimi-VL & 18.46 & 39.44 & 28.53 & 5.16 & 43.97 & 35.19 & 30.46 \\
        
        Janus-Pro & 12.40 & 10.03 & 13.24 & \best{51.89} & 34.51 & 33.33 & 28.60 \\
        
        Phi-4-multimodal-instruct & 20.11 & 27.86 & 24.54 & 16.66 & 37.56 & 0.34 & 21.39 \\     
\bottomrule
\end{tabularx}
\end{table}

% Table2 N tokens used; vs performace

\subsection{Token Efficiency and Performance Trade-off}
To better understand the computational cost associated with different models, we analyze the total number of tokens generated by each model when completing the benchmark, as well as the relationship between token usage and model performance.

\begin{figure}[htb]
  \centering
   \includegraphics[width=0.9\linewidth]{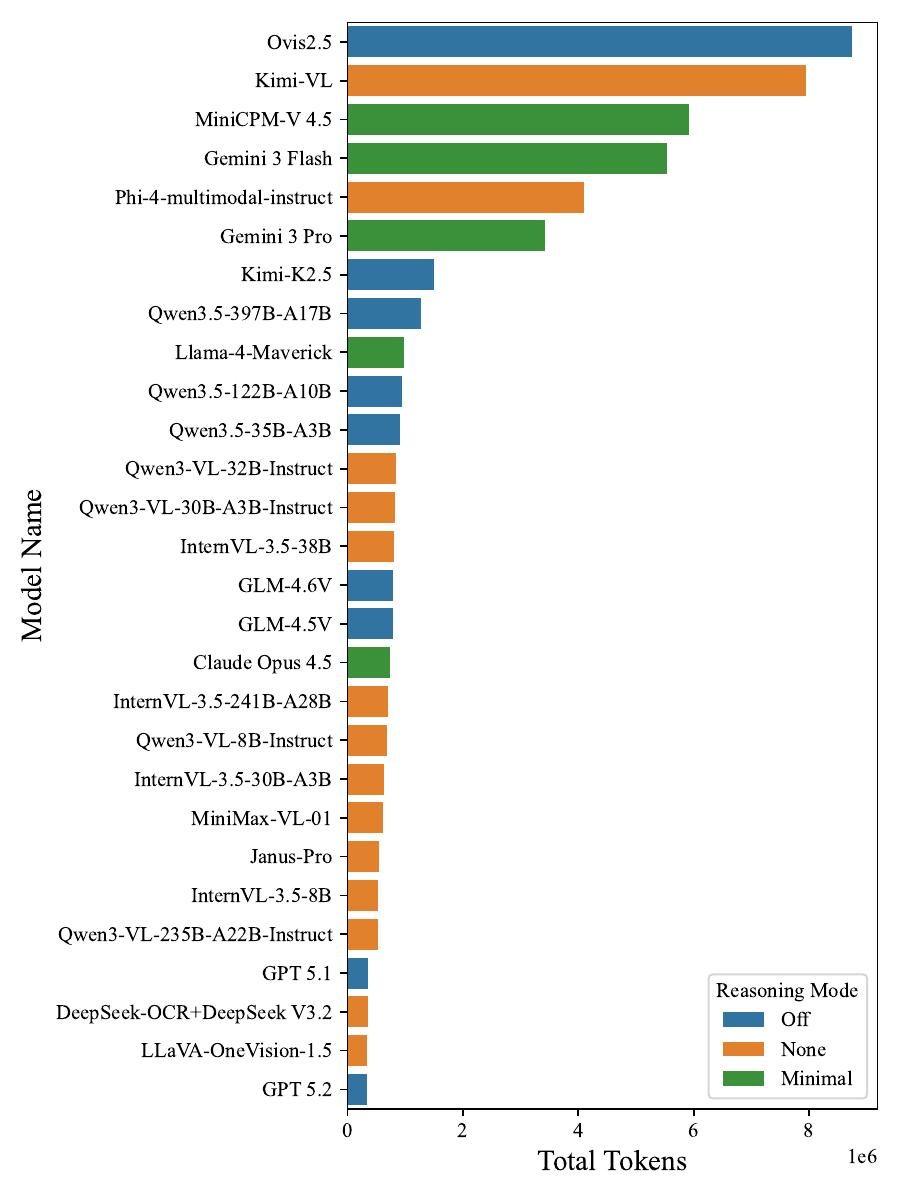}
  \caption{The total output tokens generated by each model for finishing this benchmark. The reasoning modes are defined as: a) Off: the model offers a switch for reasoning mode and our setting is "off". b) None: the model is an "instruct only" model that does not have reasoning mode. c) Minimal: the model's reasoning mode cannot be turned off, and we set the reasoning level to be "minimal".}
  \label{fig:model-tokens}
\end{figure}

Figure~\ref{fig:model-tokens} presents the total output tokens generated by each model across the entire benchmark. The results reveal substantial variation in token consumption across models, particularly between reasoning-enabled models and instruction-only models. Models operating with explicit reasoning modes tend to produce longer outputs due to intermediate reasoning traces, which can significantly increase total token usage. In contrast, instruction-only models typically generate shorter outputs focused on final answers.

\begin{figure}[ht]
  \centering
   \includegraphics[width=0.9\linewidth]{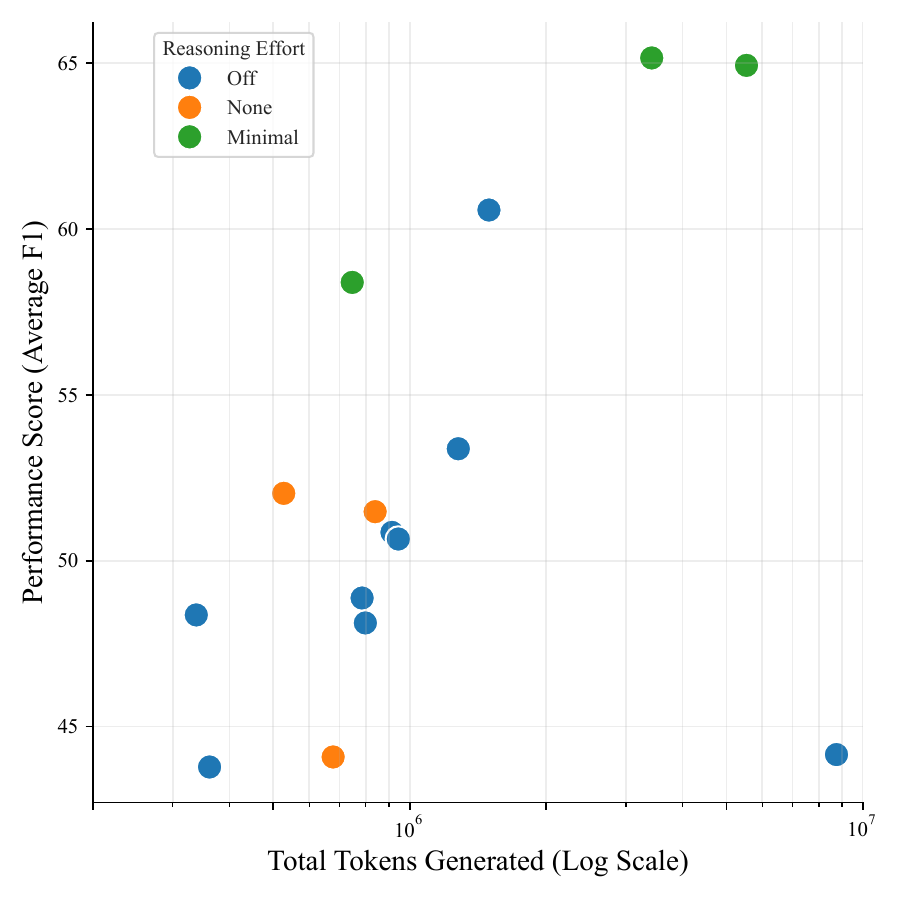}
  \caption{Tokens generated vs average F1 scores of better performing models.}
  \label{fig:token-f1}
\end{figure}

To further analyze the relationship between computational cost and model performance, we plot the average F1 score against the total tokens generated for the better-performing models (Figure~\ref{fig:token-f1}). The scatter plot suggests that higher token usage does not necessarily correspond to better performance. While some high-performing models generate large numbers of tokens due to detailed reasoning traces, several models achieve competitive performance with substantially lower token consumption.

\subsection{Human Upperbound under Robustness Challenges}
To estimate the theoretical upper bound of FCMBench under real-world robustness challenges, we conduct a human study using the same prompts and same input images as the benchmark. Specifically, we sample a total of 100 test instances from the benchmark, with approximately 10 instances per robustness challenge category (Normal Captures, Off-axis Viewpoints, Uneven Illumination, Specular Reflections, Out-of-focus, Small ROIs, Secondary Captures, Cluttered Background, Overlaid Watermarks, Cropped Captures, and Multi-doc Images). For each sampled instance, a human annotator provides a response following the original task instruction (including the same output format constraints as required by the benchmark prompts).

Human responses are evaluated against the benchmark ground truth using the same automatic evaluation script as for all model results (i.e., the same normalization and set-based exact-match F1 computation). This protocol intentionally mirrors the model evaluation pipeline, so that the resulting scores quantify a consistent human upper bound for each robustness condition.

\begin{figure}[ht]
  \centering
   \includegraphics[width=0.9\linewidth]{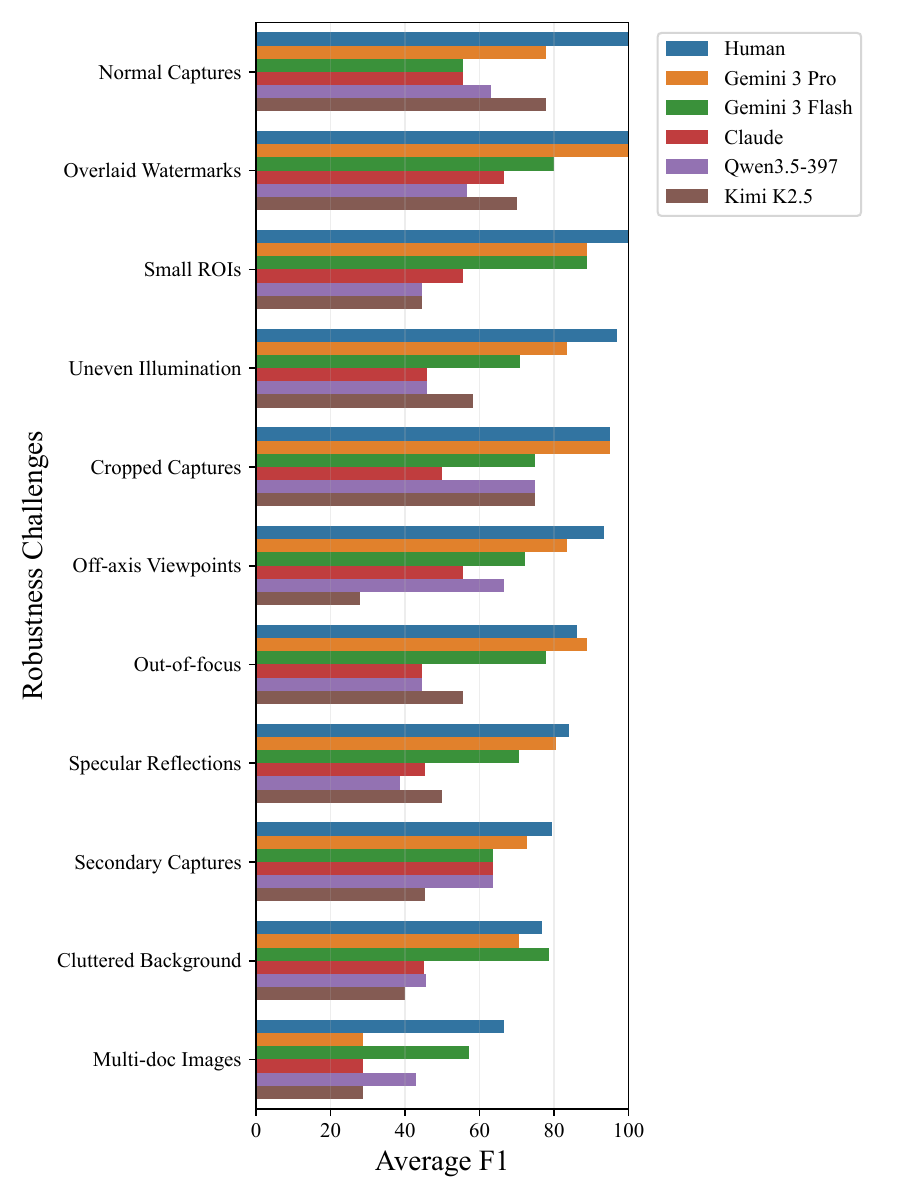}
  \caption{Tokens generated vs average F1 scores of better performing models.}
  \label{fig:human_robustness}
\end{figure}

Figure~\ref{fig:human_robustness} reports the average F1 scores of humans alongside representative top-performing models across robustness challenges. Overall, humans achieve consistently higher scores across most challenge types, providing an empirical estimate of the achievable performance ceiling under the benchmark’s strict exact-match criteria. Meanwhile, the human scores also exhibit noticeable degradation under the most severe artifacts (e.g., heavy occlusions, small ROIs, and multi-document compositions), indicating that a non-trivial fraction of errors are attributable to intrinsic information loss caused by acquisition artifacts rather than model limitations alone.

\end{document}